\documentclass[10pt,journal,compsoc]{IEEEtran}
\ifCLASSINFOpdf
\else
\fi
\usepackage{booktabs}
\usepackage{multirow}
\usepackage{times}
\usepackage{xcolor}
\usepackage{soul}
\usepackage[utf8]{inputenc}
\usepackage{epsfig}
\usepackage{graphicx}
\usepackage{amsmath}
\usepackage{amssymb}
\usepackage{url}  %Required
\usepackage{amsmath}
\usepackage{adjustbox}
\usepackage{subfigure}
\usepackage{blindtext}
\usepackage{enumerate}
\usepackage{bm}
\usepackage{dsfont}
\usepackage{array}
\usepackage[numbers]{natbib}
\usepackage[hidelinks]{hyperref}
\usepackage{balance}
\usepackage[ruled,lined,linesnumbered]{algorithm2e}
\usepackage{amsfonts}
\usepackage{setspace}
\usepackage{ragged2e}
\usepackage{lineno}
\usepackage{comment}
\usepackage{color}
\usepackage{tabularx}
\usepackage{makecell}
\usepackage{xspace}
\usepackage{colortbl}
\usepackage{multirow}
\usepackage{multicol}

\renewcommand{\raggedright}{\leftskip=0pt \rightskip=0pt plus 0cm}
\newcolumntype{L}[1]{>{\raggedright\arraybackslash}p{#1}}
\newcolumntype{C}[1]{>{\centering\arraybackslash}p{#1}}
\newcolumntype{R}[1]{>{\raggedleft\arraybackslash}p{#1}}
\newlength\savedwidth

\graphicspath{ {figures/} }

\makeatletter
\DeclareRobustCommand\onedot{\futurelet\@let@token\@onedot}
\def\@onedot{\ifx\@let@token.\else.\null\fi\xspace}
\def\eg{\emph{e.g}\onedot} 
\def\ie{\emph{i.e}\onedot} 
 
\def\etc{\emph{etc}\onedot}

\makeatother

\begin{document}
\title{Affective Image Content Analysis: Two Decades Review and New Perspectives}

%\author{
%\thanks{Copyright (c) 2013 IEEE. Personal use of this material is permitted. However, permission to use this material for any other purposes must be obtained from the IEEE by sending a request to pubs-permissions@ieee.org.}
% \thanks{author info.}
% \thanks{author info.}
% \thanks{author info.}
% \thanks{author info.}
% \thanks{author info.}
% \thanks{author info.}
% \thanks{author info.}
% \thanks{author info.}
% \thanks{author info.}
% \thanks{author info.}
% \thanks{author info.}
% \thanks{author info.}
% \thanks{author info.}
% \thanks{author info.}
% \thanks{author info.}

\author{Sicheng~Zhao,\IEEEmembership{~Senior~Member,~IEEE},~Xingxu~Yao,~Jufeng~Yang,~Guoli~Jia,~Guiguang~Ding,\\~Tat-Seng~Chua,~Bj{\"o}rn~W.~Schuller,\IEEEmembership{~Fellow,~IEEE},~Kurt~Keutzer,\IEEEmembership{~Life~Fellow,~IEEE}
\IEEEcompsocitemizethanks{\IEEEcompsocthanksitem S. Zhao and G. Ding are with BNRist, Tsinghua University, Beijing 100084, China. (e-mail: schzhao@gmail.com, dinggg@tsinghua.edu.cn).\protect
\IEEEcompsocthanksitem X. Yao, J. Yang (corresponding author), and G. Jia are with the College of Computer Science, Nankai University, China (e-mail: yxx\_hbgd@163.com, yangjufeng@nankai.edu.cn, exped1230@gmail.com).\protect
\IEEEcompsocthanksitem T.-S. Chua is with the School of Computing, National University of Singapore, Singapore (e-mail: dcscts@nus.edu.sg).\protect
\IEEEcompsocthanksitem B. W. Schuller is with the Department of Computing, Imperial College London, UK (e-mail: bjoern.schuller@imperial.ac.uk).\protect
\IEEEcompsocthanksitem K. Keutzer is with the Department of Electrical Engineering and Computer Sciences, University of California, Berkeley, USA (e-mail: keutzer@berkeley.edu).
}
%\thanks{Manuscript received March 19, 2020, revised November 29, 2020 and May 19, 2021.}
}


\markboth{IEEE Transactions on Pattern Analysis and Machine Intelligence,~Vol.~X, No.~X, June~2021}%
{Zhao \MakeLowercase{\textit{et al.}}: Affective Image Content Analysis}

\IEEEtitleabstractindextext{%
\begin{abstract}
Images can convey rich semantics and induce various emotions in viewers. Recently, with the rapid advancement of emotional intelligence and the explosive growth of visual data, extensive research efforts have been dedicated to affective image content analysis (AICA). In this survey, we will comprehensively review the development of AICA in the recent two decades, especially focusing on the state-of-the-art methods with respect to three main challenges -- the affective gap, perception subjectivity, and label noise and absence. We begin with an introduction to the key emotion representation models that have been widely employed in AICA and description of available datasets for performing evaluation with quantitative comparison of label noise and dataset bias. We then summarize and compare the representative approaches on (1) emotion feature extraction, including both handcrafted and deep features, (2) learning methods on dominant emotion recognition, personalized emotion prediction,  emotion distribution learning, and learning from noisy data or few labels, and (3) AICA based applications. Finally, we discuss some challenges and promising research directions in the future, such as image content and context understanding, group emotion clustering, and viewer-image interaction.
\end{abstract}

}

\maketitle

\IEEEdisplaynontitleabstractindextext

\IEEEpeerreviewmaketitle

\IEEEraisesectionheading{\section{Introduction}
\label{sec:Introduction}}

\IEEEPARstart{I}{n} the book ``The Society of Mind''~\cite{minsky1988society}, Minsky (a Turing Award winner in 1970) claimed that ``\textit{The question is not whether intelligent machines can have any emotions, but whether machines can be intelligent without emotions.}'' Although emotions play a vitally important role in machine and artificial intelligence, much less attention has been paid to affective computing than objective semantic understanding, such as object classification in computer vision. The rapid development of artificial intelligence has made remarkable success in semantic understanding and raised higher demand to emotional interaction. For example, the companion robots that can recognize and express emotions can provide more harmonious companionship for human beings, especially the elderly and single children. To have human-like emotions, machines should first understand how humans express emotions through multiple channels, such as speech, gesture, facial expression, and physiological signals~\cite{d2015review}. While other signals can be easily suppressed or masked, physiological signals that are controlled by the sympathetic nervous systems are independent of
%BS:
humans'
will and thus provide more reliable information. However, to capture accurate physiological signals is quite difficult and impractical, as it requires special types of wearable sensors. On the other hand, the recent convenient access of cameras in mobile devices and wide popularity of social networks (such as Twitter, Flickr, and Weibo) have enabled people to habitually share their experiences and express their opinions online using images and videos together with text~\cite{zhao2020endtoend}. Recognizing the affective content of this large volume of multimedia data provides an alternate way to understand users' behaviors and emotions.

%\IEEEPARstart{I}{n} the book ``The Society of Mind''~\cite{minsky1988society}, Minsky (a Turing Award winner in 1970) claimed that ``\textit{The question is not whether intelligent machines can have any emotions, but whether machines can be intelligent without emotions.}'' To have human-like emotions, machines should first understand how humans express emotions through multiple channels, such as speech, gesture, facial expression, and physiological signals~\cite{d2015review}. While other signals can be easily suppressed or masked, physiological signals that are controlled by the sympathetic nervous systems are independent of humans’ will and thus provide more reliable information. However, capturing accurate physiological signals is quite difficult and unpractical, which requires types of wearable sensors. On the other hand, the recent convenient access of cameras in mobile devices and wide popularity of social networks (such as Twitter, Flickr) have enabled people habitually share their experiences and express their opinions online using images and videos together with text. Recognizing the affective content of this large volume of multimedia data can help to understand users' behaviors and emotions.

%With the convenient access of cameras in mobile devices and the immense popularity of social networks

%Tremendous advances have been made in artificial intelligence with the advent of deep learning.

\begin{figure}[!t]
\begin{center}
\includegraphics[width=0.95\linewidth]{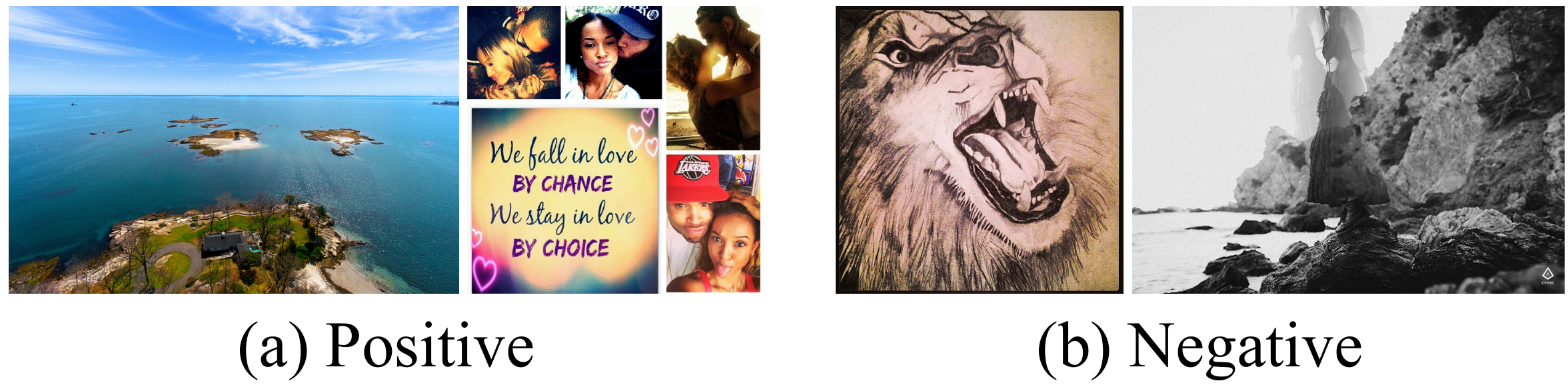}
\caption{Examples of relevance and importance of AICA to infer humans' emotional status. Images are from the FI dataset~\cite{you2016building}.}
\label{fig:AICAExamples}
\end{center}
\end{figure}

As we know, ``a picture is worth a thousand words'', which indicates that images can convey rich semantics. Different from existing research on analyzing the perceptual aspects of images, such as object detection and semantic segmentation, affective image content analysis (AICA) focuses on understanding the semantics at a higher level --
%BS:
the
cognitive level, \ie understanding the emotions that can be induced by the images in viewers, which is more challenging.
The automatic inference of humans' emotional status using AICA can help to evaluate their psychological health, discover affective anomaly, and prevent extreme behaviors to themselves and even to the whole society. For example, in Fig.~\ref{fig:AICAExamples}, the users posting images (b) are more likely to have negative emotions than the users posting images (a).

%the users posting images (b) are more likely to have negative emotions and take revenge on society to express their dissatisfaction than the users posting images (a).

\subsection{Main Goals and Challenges}
\label{ssec:Challenges}

\textbf{Main Goals.} Given an input image, AICA mainly aims to (1) recognize the emotions that can be induced to specific viewers or to the majority (Based on psychology, the emotions might be represented in different models, \eg categorical or dimensional. Please see Section~\ref{sec:EmotionModels} for details.), (2) analyze what stimuli contained in the image evoke such emotion (\eg specific objects or color combinations), and (3) apply the recognized emotions to different real-world applications to improve the ability of emotional intelligence.

\begin{figure}[!t]
\begin{center}
\subfigure[Affective gap overview]{
\includegraphics[width=0.95\linewidth]{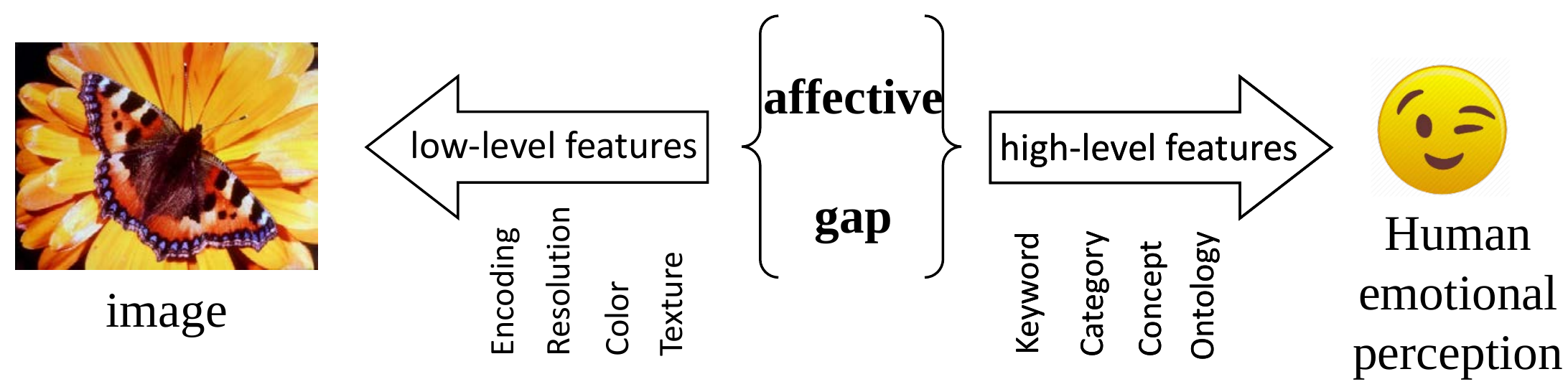}
}
\subfigure[Affective gap examples]{
\includegraphics[width=0.95\linewidth]{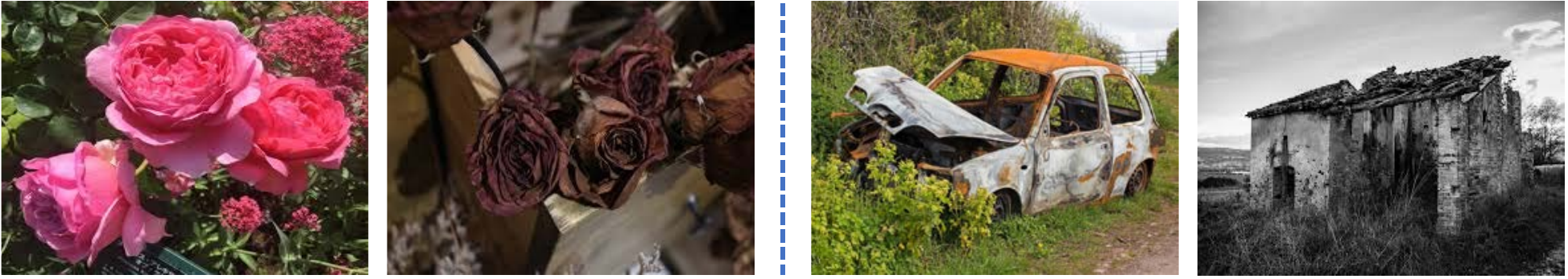}
}
\caption{Illustration of the affective gap. (a) Overview: the commonly extracted low-level features cannot well represent high-level emotions. (b) Examples: the first pair of images have a similar object (rose) but evoke different emotions, while the second pair of images exhibit entirely different content (car vs. house) but evoke similar emotions.}
\label{fig:AffectiveGap}
\end{center}
\end{figure}

\noindent\textbf{Challenges. (1) Affective Gap.}
Similar to the semantic gap in computer vision,
%BS:
the
affective gap is one main challenge for AICA, which can be defined as ``the lack of coincidence between the features and the expected affective state in which the user is brought by perceiving the signal''~\cite{hanjalic2006extracting}, as shown in Fig.~\ref{fig:AffectiveGap}.

\begin{figure}[!t]
\begin{center}
\includegraphics[width=0.95\linewidth]{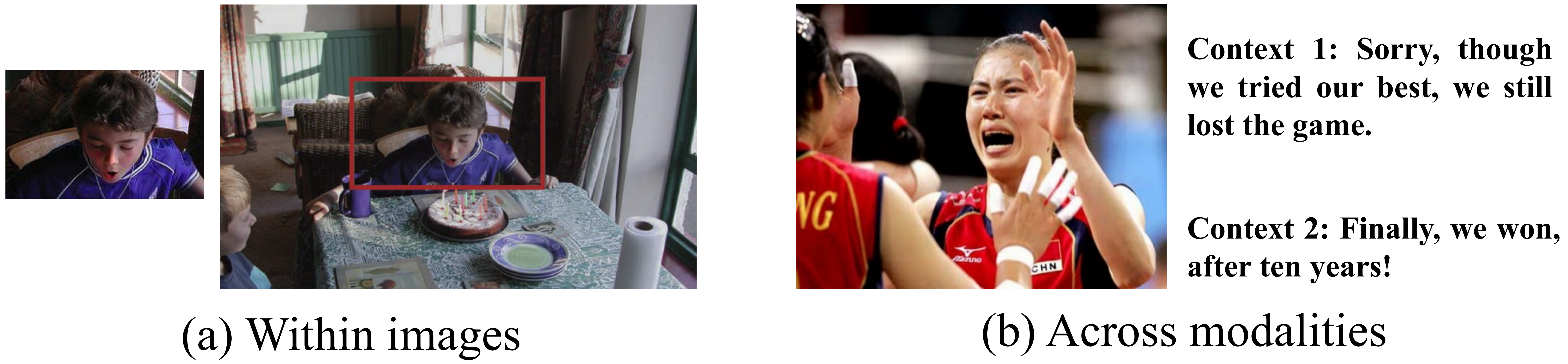}
\caption{The context information also plays an important role in AICA. (a) The image without and with the detailed scene context evoke different emotions (surprise vs. happy). (b) The textual contexts can also influence the emotion perception of the same image (sad vs. excited).
}
\label{fig:ImageContext}
\end{center}
\end{figure}

%BS introduce GIST spelt out?
%SZ: The “gist” is an abstract representation of the scene that spontaneously activates memory representations of scene categories (a city, a mountain, etc.)
% We use Gist instead.
To bridge the affective gap, researchers primarily focus on extracting discriminative features that can better distinguish the difference among different emotions, ranging from hand-crafted features like Gabor~\cite{yanulevskaya2008emotional}, Gist~\cite{patterson2012sun}, artistic elements~\cite{machajdik2010affective}, artistic principles~\cite{zhao2014exploring}, and adjective noun pairs (ANPs)~\cite{borth2013large} to deep ones like convolutional neural networks (CNNs)~\cite{chen2015learning,you2016building} and regions~\cite{yang2018weakly}. Based on the assumption that different viewers can reach a consensus on the perceived emotions of images, these AICA methods mainly assign an image with the dominant (average) emotion category (DEC). This task can be performed as a traditional single-label learning problem.

Besides extracting visual features, incorporating available context information can also contribute to the AICA task~\cite{kosti2020context}, as shown in Fig.~\ref{fig:ImageContext}. The same image under different contexts may evoke different emotions. For example, in Fig.~\ref{fig:ImageContext} (a), if we just see the kid, we may feel surprise based on his expression; but with the context that the kid is blowing the candles to celebrate his birthday, it is more likely to make us feel happy. In Fig.~\ref{fig:ImageContext} (b), if we see a volleyball player crying, we may feel sad; but if there is a comment for the image, ``Finally, we won, after ten years!'', we, especially the volleyball amateurs of the team, may feel excited.

\begin{figure}[!t]
\begin{center}
\includegraphics[width=0.95\linewidth]{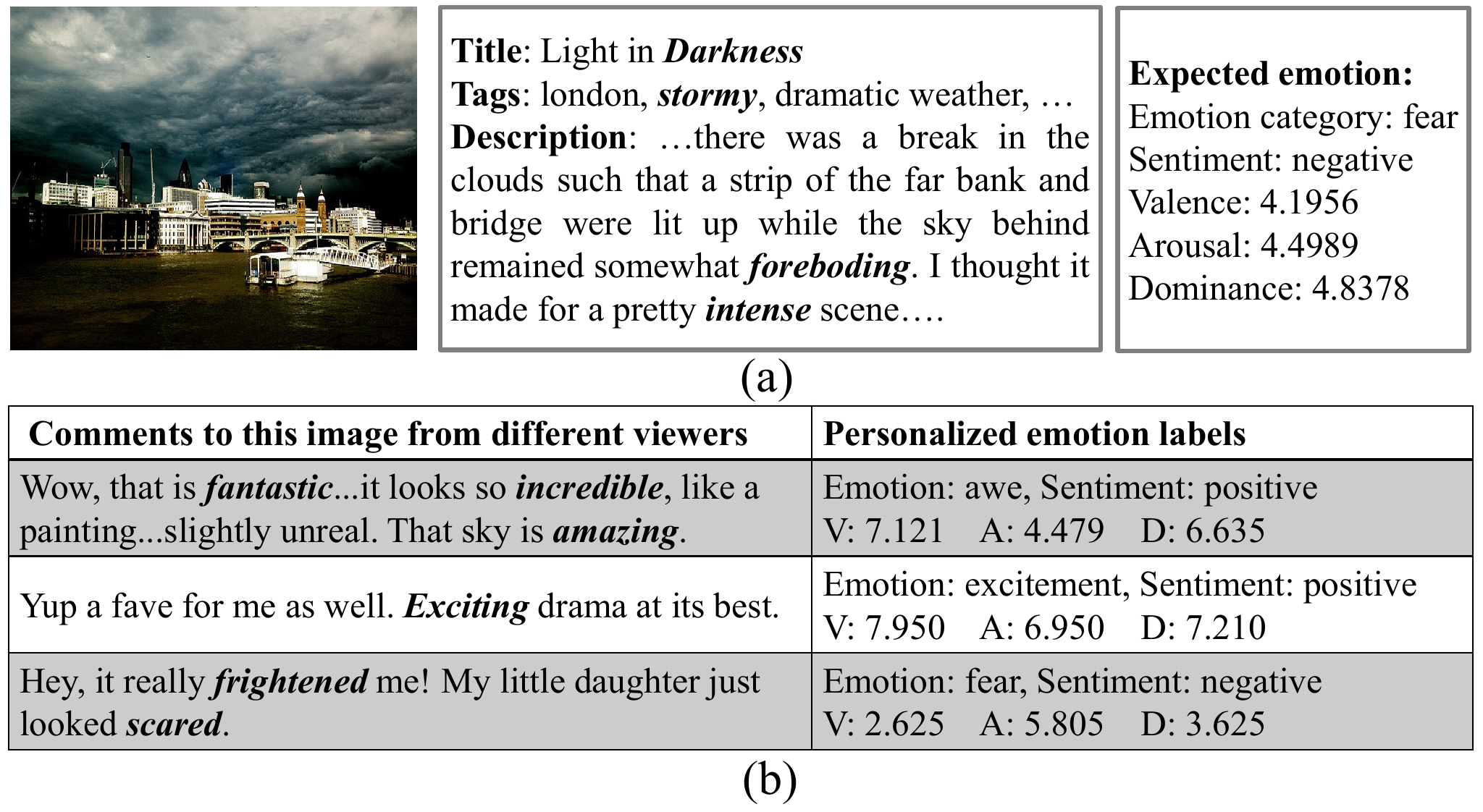}
\caption{Illustration of the perception subjectivity~\protect\cite{zhao2017continuous}. For the original image (a) uploaded to Flickr, different viewers may have different emotion perceptions (b). The emotion labels are obtained using the keywords in italic based on the comments from these viewers.
%(a) The original image uploaded to Flickr. (b) Personalized emotion perceptions with labels obtained using the keywords in red based on the comments from different viewers.
}
\label{fig:SubjectivityExamples}
\end{center}
\end{figure}

\noindent\textbf{(2) Perception Subjectivity.}
Different viewers may have totally different emotional reactions to the same image, which is caused by many personal and contextual factors, such as the cultural background, personality and social context~\cite{peng2015mixed,zhao2016predicting,yang2017learning}. For example, for the ``Light in Darkness'' image in Fig.~\ref{fig:SubjectivityExamples}(a), viewers who are interested in capturing natural
%BS:
phenomena
are probably excited to see this spectacle, while the viewers who are scared of thunder and storm might feel fear. This fact causes the so-called subjective perception
problem. Therefore, for this highly subjective variable, simply predicting the DEC is insufficient, since it cannot well reflect the difference among different viewers.

To tackle the subjectivity issue, we can conduct two kinds of AICA tasks~\cite{zhao2016predicting}: for each viewer, we can predict personalized emotion perceptions; for each image, we can assign multiple emotion labels. For the latter one, we can employ multi-label learning methods, which associate one image with multiple emotion labels. However, since the importance or extent of different emotion labels is in fact unequal, emotion distribution learning would make more sense, which aims to learn the degree to which each emotion describes the image~\cite{yang2017learning}.

\begin{figure}[!t]
\begin{center}
\subfigure[Domain shift]{
\includegraphics[width=0.95\linewidth]{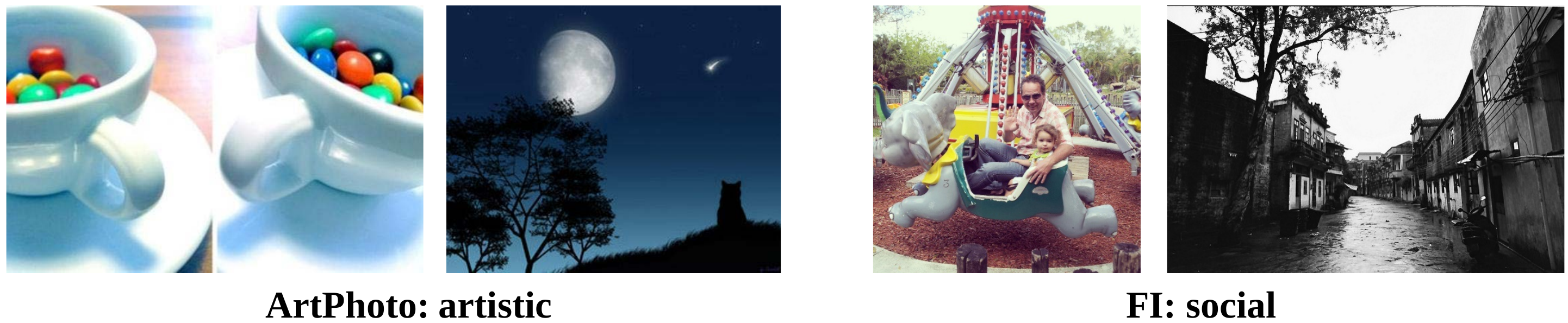}
}
\subfigure[Performance evaluation]{
\includegraphics[width=0.7\linewidth]{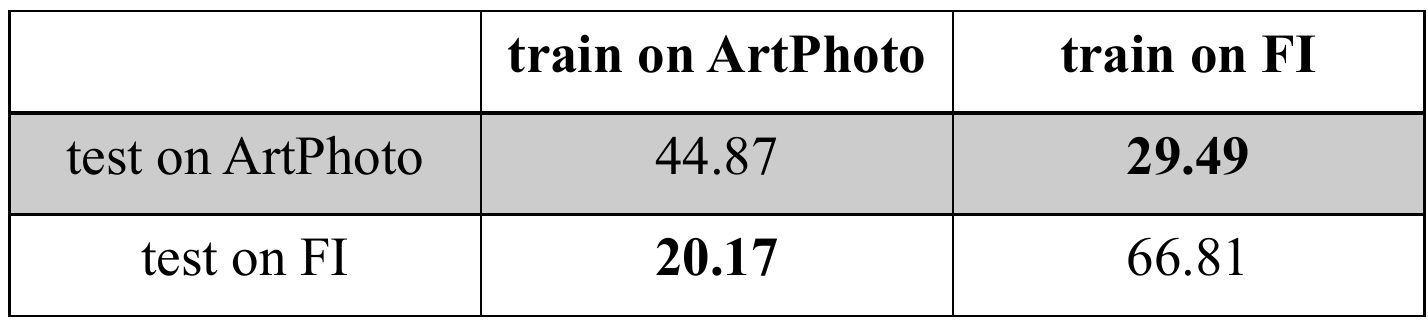}
}
\caption{Illustration of domain shift. (a) The images from ArtPhoto~\cite{machajdik2010affective} and FI~\cite{you2016building} datasets have different styles: artistic vs.\ social. (b) The emotion classification performance (\%) significantly drops if the trained dataset is different from the tested dataset on both ArtPhoto and FI datasets by fine-tuning the ResNet-101 model~\cite{he2016deep}.}
\label{fig:DomainShift}
\end{center}
\end{figure}

\begin{figure}[!t]
\begin{center}
\includegraphics[width=0.98\linewidth]{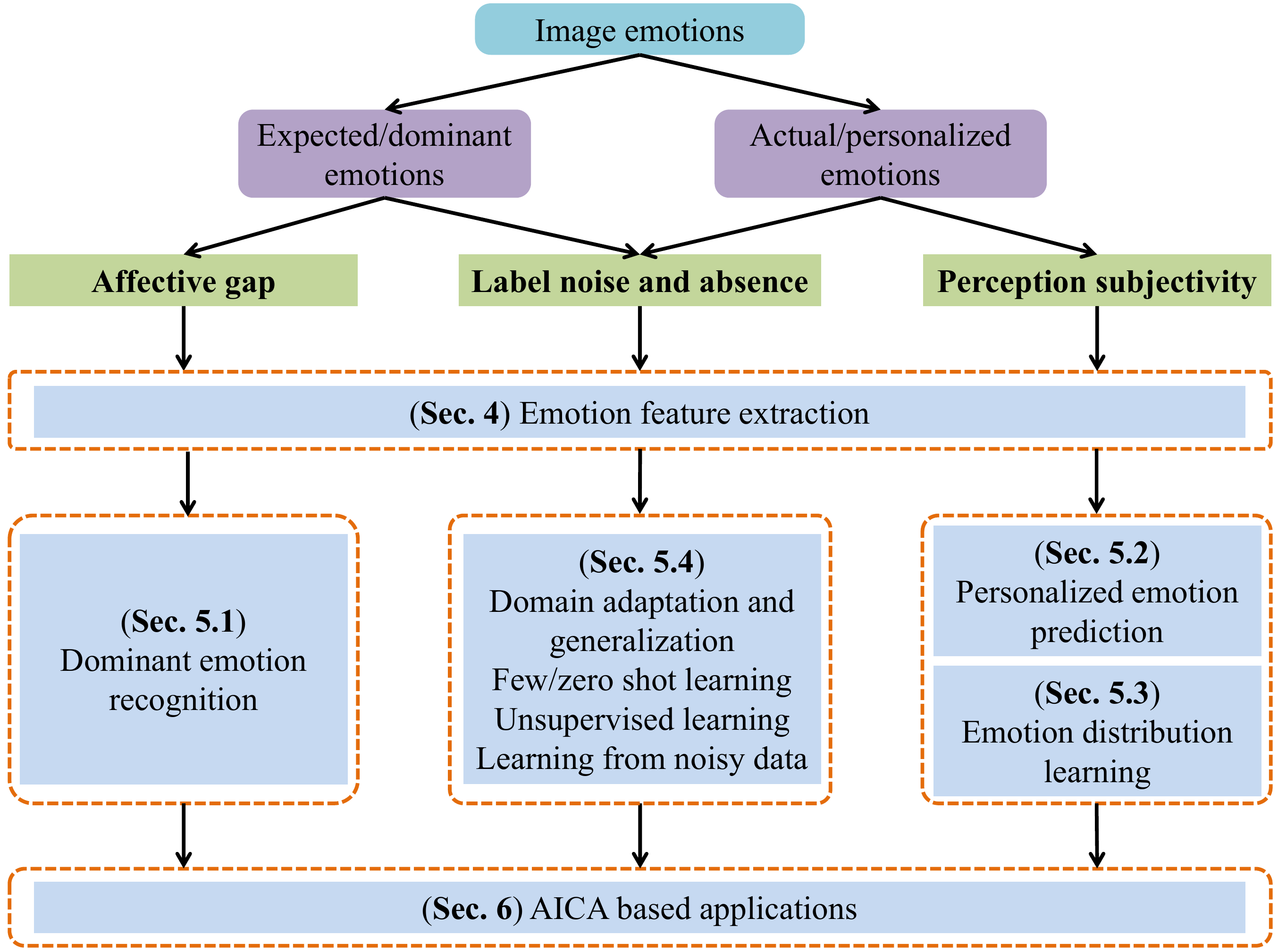}
\caption{Organization of different technical components in this survey.
}
\label{fig:OverallFramework}
\end{center}
\end{figure}

\noindent\textbf{(3) Label Noise and Absence.}
Recent methods on AICA based on deep learning, especially CNN, have achieved promising results. However, training these models requires large-scale labeled data, which is prohibitively expensive and time-consuming to obtain, not only because labeling the emotions in ground-truth generation is highly inconsistent, but also because in some cases like artistic works only experts are able to provide reliable labels. In real-world applications, there might be only few or even no labeled emotion data. How to deal with this situation is significantly worth investigating. Unsupervised/weakly supervised learning and few/zero shot learning are two interesting directions.

One possible solution is to leverage the unlimited amount of web images with associated tags as labels~\cite{wei2020learning}. However, such tags can be incomplete and noisy. An image might be associated with tags that are unrelated or remotely related. How to learn from noisily labeled images is the main challenge. Imposing some constraints for visual representation based on the semantic correlations between image and text is one direct solution. Firstly learning text models and embeddings in unsupervised or semi-supervised manners and then denoising the %BS:
keyword
labels can help to ``clean'' the label noise.

Furthermore, if we have sufficient labeled data in one domain, such as abstract paintings, how can we effectively transfer the well-trained models to another unlabeled or sparsely labeled domain? Because of the presence of \emph{domain shift} or \emph{dataset bias}~\cite{torralba2011unbiased,zhao2020review}, direct transfer often results in poor performance, as shown in Fig.~\ref{fig:DomainShift}. Specifically, \citeauthor{panda2018contemplating}~\cite{panda2018contemplating} classified the dataset bias in AICA into two categories. One is positive set bias. Due to the lack of diversity in visual concepts for each emotion category (\eg amusement) in the source domain, the models learned based on such data are easily to memorize all its idiosyncrasies and lose the ability to generalize to the target domain. The other is negative set bias. The rest of the dataset (\eg the data coming from other categories excluding amusement) in the source domain does not well represent the rest of the visual world. For example, some of the negative samples from the target domain are confused with the positive samples in the source domain. As a result, the learned classifiers might be overconfident. Domain adaptation and domain generalization might help to address this issue.

\subsection{Organization of This Survey}
In this survey, we concentrate on reviewing the state-of-the-art methods on AICA and outlining research trends. First, we introduce the brief history in Section~\ref{ssec:History} and its comparison with other related topics in Section~\ref{ssec:Comparison}. Second, we describe the widely-used emotion representation models in Section~\ref{sec:EmotionModels}. Third, we summarize the available datasets for performing AICA evaluation in Section~\ref{sec:Datasets} and quantitatively compare the label noise and dataset bias. Fourth, based on the main goals and challenges in Section~\ref{ssec:Challenges}, we summarize and compare the representative approaches on emotion feature extraction, learning methods (for dominant emotion recognition, personalized emotion prediction, emotion distribution learning, and learning from noisy data or few labels), and AICA based applications in Sections~\ref{sec:Features}, \ref{sec:Methods}, and \ref{sec:Applications}, respectively, as shown in Fig.~\ref{fig:OverallFramework}. Finally, we discuss potential research directions to pursue in Section~\ref{sec:FutureDirections}.

\begin{figure*}[!t]
\begin{center}
\includegraphics[width=0.9\linewidth]{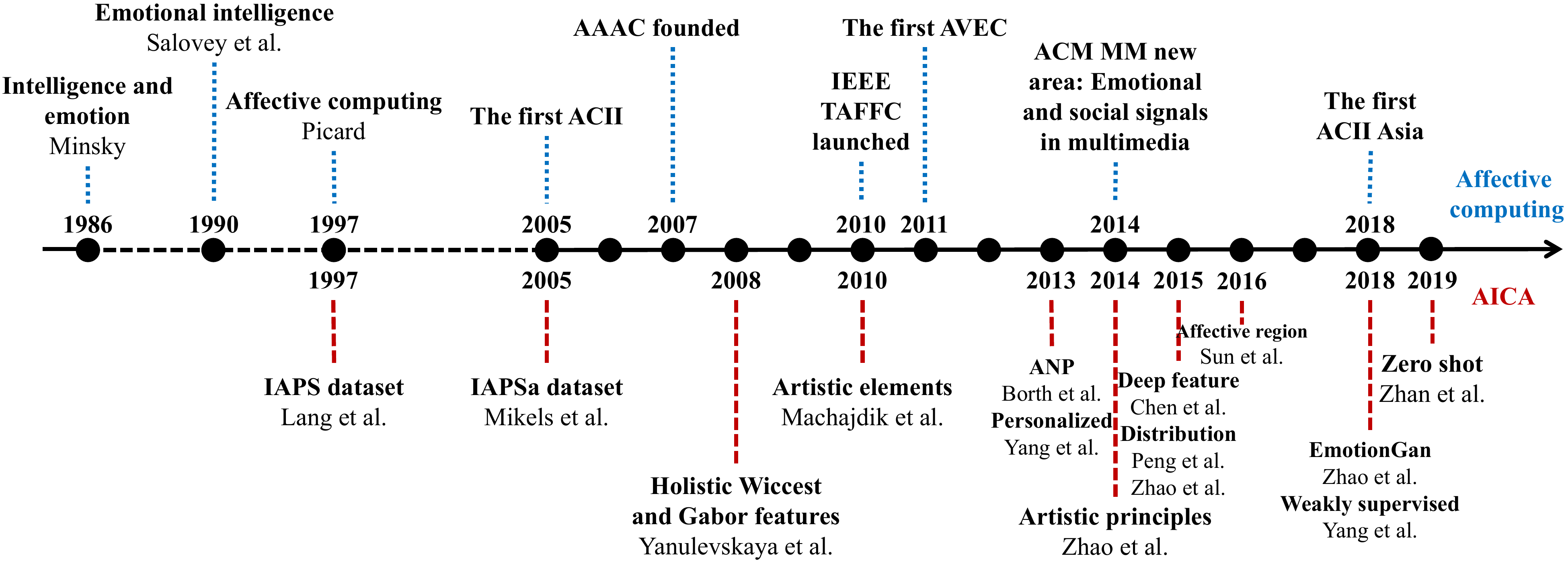}
\caption{Milestones in both general affective computing (above line, blue) and affective image content analysis (below line, red).}
\vspace{-12pt}
\label{fig:History}
\end{center}
\end{figure*}
%The names not in bold under each event are corresponding authors.

\subsection{Brief History}
\label{ssec:History}

\textbf{Affective Computing.}
%BS: Start
Before affective computing was known by this term, early first works include a 1978 filed patent on an analyzer for determining the emotion of a speaker speech~\cite{williamson1979speech}, and scientific papers on generation of affect in synthesized speech in 1990~\cite{cahn1990generation}, or the recognition of facial expressions by neural networks in 1992~\cite{kobayashi1992recognition}.
%BS: End

Since Minsky proposed the emotion recognition problem of intelligent machines~\cite{minsky1988society}, much attention has been paid to emotion related research, such as the definition of emotional intelligence~\cite{salovey1990emotional}. In 1997, Picard first proposed the concept of affective computing~\cite{picard1997affective}: ``affective computing is computing that relates to, arises from, or deliberately influences emotion or other affective phenomena''. Some influential events include: the first International Conference on Affective Computing and Intelligent Interaction (ACII) by IEEE and AAAI in 2005, the foundation of
the Association for the Advancement of Affective Computing (AAAC) in 2007
%BS:
(originally named HUMAINE Association),
%BS: Start
the first ever public `emotion challenge' held at Interspeech 2009,
%BS: End
the launch of the IEEE Transactions on Affective Computing (TAFFC) in 2010, the first International Audio/Visual Emotion Challenge and Workshop (AVEC) in 2011, the proposal of the Emotional and Social Signals in Multimedia area in ACM Multimedia 2014, and the first ACII Asia in 2018, \textit{etc.}
%the first international challenge to also feature physiology (AVEC) in 2015,

\noindent\textbf{Affective Image Content Analysis.}
The development of AICA begins in the psychology and behavior research, such as the International Affective Picture System (IAPS)~\cite{lang1997international,mikels2005emotional}, to investigate the relation between visual stimuli and emotion. One of the first emotion recognition methods is based on low-level holistic Wiccest and Gabor features~\cite{yanulevskaya2008emotional}. Since then,
%BS:
several representative hand-crafted features have been designed, such as the low-level artistic elements~\cite{machajdik2010affective}, mid-level artistic principles~\cite{zhao2014exploring}, and high-level Adjective Noun Pairs (ANPs)~\cite{borth2013large}. In 2014, transfer learning is conducted from a CNN in which parameters are pre-trained by large-scale data~\cite{xu5731visual}. To tackle the subjectivity challenge, both personalized emotion prediction~\cite{yang2013user,zhao2016predicting} and emotion distribution learning~\cite{peng2015mixed,zhao2015predicting,zhao2015continuous,liu2018low} are considered. More recently, domain adaptation~\cite{zhao2018emotiongan,zhao2019cycleemotiongan,zhao2021emotional} and zero-shot learning~\cite{zhan2019zero} are studied for the label absence challenge. The representative milestones in both general affective computing and AICA are summarized in Fig.~\ref{fig:History}.

\subsection{Comparison with Other Related Topics}
\label{ssec:Comparison}

\textbf{Comparison with Affective Computing of Other Modalities.} Affective content analysis has also been widely studied in other modalities, such as
text~\cite{giachanou2016like,zhang2018deep},
%BS:
speech acoustics~\cite{el2011survey,schuller2018speech} and linguistics~\cite{metze2010emotion}, music~\cite{schuller2010mister,yang2012machine},
%BS
sound~\cite{schuller2012automatic},
facial expression~\cite{pantic2006dynamics,sariyanidi2014automatic,li2018deep,hassan2019automatic}, video~\cite{wang2015video,zhao2020end}, physiological signals~\cite{alarcao2017emotions,keren2017end,zhao2019personalized}, and multi-modal data~\cite{schuller2002multimodal,soleymani2017survey,poria2017review,zhao2019affective}. Although the employed emotion models and learning methods are similar, there is
%BS
a
clear difference between affective computing of images and other modalities, especially the extracted features to represent emotions. While the surveys on other modalities are well-conducted, there is no comprehensive survey on AICA. As
%BS:
an
image is an important channel to express emotions, we believe an in-depth analysis of AICA could boost the development of affective computing. One preliminary version on this survey was previously introduced in our IJCAI 2018 conference paper~\cite{zhao2018affective}. As compared to the conference paper, this journal version has
the following five aspects of extensive enhancements. First, the detailed challenges and a brief history are incorporated. Second, we summarize and compare more representative works on emotion models, available datasets, emotion features, and learning methods. Third, we conduct extensive experiments to fairly compare the effectiveness of different AICA methods. Fourth, we add some AICA-based applications. Finally, we discuss more potential research directions.

\noindent\textbf{Comparison with Computer Vision.} The task of AICA is often composed of three steps: human annotation, visual feature extraction, and learning of mapping between visual features and perceived emotions~\cite{zhao2017approximating}. Although the three steps seem to be very similar to computer vision (CV, the third step is a mapping learning between visual features and image labels, such as
%BS:
an
object), there are significant differences between AICA and CV.%, as shown in Table~\ref{tab:CV_AICA}.
Take object classification and emotion classification for instance. (1) Even
%BS:
if
the semantic gap is bridged in object classification, there still exists
%BS:
an
affective gap. For example,
%BS:
an image with a lovely dog and
%BS:
an image with a barking dog can evoke different emotions.
%Therefore, AICA requires more fine-grained object recognition with subjective labeling.
(2) Object is an objective concept (both a lovely dog and a barking dog are dogs), while emotion is a relatively subjective concept (happy and fear for the two images). (3) Correspondingly, object classification belongs to the perceptual aspects of images, while AICA focuses on the cognitive level. Object classification is mainly studied by the CV community, while AICA is an interdisciplinary task requiring
psychology, cognitive science, multimedia, and machine learning, \textit{etc.}

\begin{table}[!t]
\centering\footnotesize
\caption{Representative emotion models employed in AICA.}
\begin{tabular}
%{C{1cm}<{\centering}  C{0.8cm}<{\centering}  C{0.81cm}<{\centering}  p{8cm}<{\centering}}
{cccp{5cm}<{\centering}}
\toprule
\textbf{Model} & \textbf{Ref} & \textbf{Type} & \multicolumn{1}{c}{\textbf{Emotion states/dimensions}} \\
\hline
Ekman & \cite{ekman1992argument}  & CES & happiness, sadness, anger, disgust, fear, surprise\\
Mikels & \cite{mikels2005emotional} & CES & amusement, anger, awe, contentment, disgust, excitement, fear, sadness\\
Plutchik & \cite{plutchik1980emotion} & CES & ($\times$ 3 scales) anger, anticipation, disgust, joy, sadness, surprise, fear, trust\\
Parrott & \cite{parrott2001emotions} & CES & a tree hierarchical grouping with primary, secondary and tertiary emotion categories\\
Sentiment & & CES & positive, negative, (and neutral)\\
VA(D) & \cite{schlosberg1954three} & DES & valence-arousal(-dominance)\\
ATW & \cite{lee2011fuzzy} &DES & activity-temperature-weight\\
\bottomrule
\end{tabular}
\label{tab:EmotionModels}
\end{table}

%\section{Emotion Models from Cognitive Science Community}
\section{Emotion Models from Psychology}
\label{sec:EmotionModels}
In psychology, there are several different affective computing related concepts, such as emotion, affect, mood, %SZ: remove etc.
and sentiment. Discussing the difference or correlation of these concepts is out of the scope of this survey. Interested readers can refer to~\cite{munezero2014they} for more details. Appraisal theory is well-known for explaining the development of emotional experience~\cite{scherer2001appraisal}. It accounts for individual variability in emotional reactions to the same stimulus. According to the Ortony, Clore and Collins (OCC) model~\cite{ortony1988cognitive}, the emergence of emotions originates from the cognitive evaluation or appraisal of stimuli in terms of events, agents, and objects. How individuals actually perceive and interpret the stimuli determines how emotions might emerge.

Psychologists mainly employ two kinds of emotion representation models to measure emotion: categorical emotion states (CES) and dimensional emotion space (DES), as shown in Table~\ref{tab:EmotionModels}. CES models classify emotions into a few basic categories. The simplest CES model is binary \emph{positive} and \emph{negative}  (polarity)~\cite{zhao2018personality,zhao2019personalized}. In such cases, ``emotion'' is often called ``sentiment'', which sometimes also includes \emph{neutral}. Since sentiment is too coarse-grained, some relatively fine-grained emotion models are designed, such as Ekman's six emotions (anger, disgust, fear, happiness, sadness, surprise)~\cite{ekman1992argument} and Mikels's eight emotions (amusement, anger, awe, contentment, disgust, excitement, fear, and sadness)~\cite{mikels2005emotional}. With the development of psychological
theories, categorical emotions are becoming increasingly diverse and fine-grained
%BS:
such as by also considering social emotions~\cite{gunes201716}.
Besides the eight basic emotion categories (anger, anticipation, disgust, fear, joy, sadness, surprise, trust), Plutchik~\cite{plutchik1980emotion} organized each of them into 3 intensities which thus provides a richer set. For example, the 3 intensities of joy and fear are ecstasy$\rightarrow$joy$\rightarrow$serenity and terror$\rightarrow$fear$\rightarrow$apprehension, respectively. Another representative CES model is Parrott's tree hierarchical grouping~\cite{parrott2001emotions}, which represents emotions with primary, secondary and tertiary categories. For example, a three-level emotion hierarchy is designed as two
basic categories (positive and negative) at level-1, six categories (anger, fear, joy,
love, sadness, and surprise) at level-2 and 25 fine-grained emotion categories at level-3.

Although CES models are easy for users to understand, limited emotion categories cannot well reflect the complexity and subtlety of emotions. Further, psychologists have not reached a consensus on how many discrete emotion categories should be included. Differently, DES models employ continuous
%BS: (note that AVEC 2011 used 4D!)
2D, 3D, or higher dimensional
Cartesian space to represent emotions, such as valence-arousal-dominance (VAD) \cite{schlosberg1954three}
%BS:
and potentially added intensity, novelty,
%SZ: arousal represents the intensity of emotions, so remove intensity here?
%intensity,
or others, and activity-temperature-weight~\cite{lee2011fuzzy}. VAD is the most widely used DES model~\cite{gunes2013categorical},
%BS:
where valence represents the pleasantness ranging from positive to negative, arousal represents the intensity of emotion ranging from excited to calm, and dominance represents the degree of control ranging from controlled to in control. In practice, dominance is difficult to measure and is often omitted, leading to the commonly used
%BS:
two-dimensional
VA space~\cite{hanjalic2006extracting}. Theoretically, every emotion can be represented as a coordinate point in the Cartesian space. However, the absolute continuous values are difficult for users to distinguish, which constraints the employment of DES models.
%where V is easier to recognize in AICA than A

\begin{table}
\centering\footnotesize
\caption{Comparison between CES and DES.}
\begin{tabular}{c|cc}
\toprule
           & CES             & DES              \\
\hline
understandability        &  easy   & difficult     \\
describability   & limited      &  unlimited        \\
perspective      & qualitative      & quantitative         \\
examples & Mikels, Plutchik & VAD \\
granularity & coarse-grained & fine-grained \\
AICA tasks & classification, retrieval & regression, retrieval \\
\bottomrule
\end{tabular}
\label{tab:CES_DES}
\end{table}

The comparison between CES and DES is shown in Table~\ref{tab:CES_DES}.
Compared to CES, DES is able to represent finer-grained and more comprehensive emotions, which reflects their difference on granularity and describability, respectively.
Further, instead of being independent from each other, they are actually related.
The relationship between CES and DES and the transformation from one to the other are studied in~\cite{sun2009improved, alarcao2018identifying}. For example, positive valence relates to a happy state, while negative valence relates to a sad or angry state; a relaxed state relates to low arousal, while anger relates to high arousal.
%BS:
To further distinguish for example anger and fear (both negative valence, but high arousal), one needs dominance (high for anger, but low for fear).
%BS End
CES and DES are mainly employed in classification and regression tasks, respectively, with discrete and continuous emotion labels. As a result, the employed learning models are different. For the affective image retrieval task, both models can be employed with different emotion distance measurements (\eg Mikels' emotion wheel~\cite{zhao2016predicting} for CES and Euclidean distance for DES). If we discretize DES into several constant values, we can also use it for classification~\cite{lee2011fuzzy}. We can consider easing DEC comprehension difficulties in raters by ranking based labeling.

\begin{table*}[!t]
	\centering\scriptsize
	\caption{Released datasets for AICA, where `\# Images' and `\# Annotators' represent the total number of images and annotators (f: female, m: male).}
	\begin{tabular}
		{l l r l r l l}
		\toprule
		\textbf{Dataset} & \textbf{Ref} & \textbf{\# Images} & \textbf{Type} & \textbf{\# Annotators}& \textbf{Emotion model} &  \textbf{Label detail}\\
		\hline
		IAPS & \cite{lang1997international}  & 1,182 & natural & $\approx$100 (half f) & VAD & empirically derived mean and standard deviation\\
		IAPSa & \cite{mikels2005emotional} & 390 & natural & 20 (10f,10m) & Mikels & at least one emotion category for each image\\
		Abstract & \cite{machajdik2010affective} & 280 & abstract & $\approx$230 & Mikels & the detailed votes of all emotions for each image\\
		ArtPhoto & \cite{machajdik2010affective} & 806 & artistic & -- & Mikels & one DEC for each image\\
		GAPED & \cite{dan2011geneva} & 730 & natural & 60  & Sentiment, VA & one DEC and average VA values for each image\\
		MART & \cite{alameda2016recognizing} & 500 & abstract & 25 (11f,14m) & Sentiment & one DEC for each image\\
		devArt & \cite{alameda2016recognizing} & 500 & abstract & 60 (27f,33m) & Sentiment & one DEC for each image\\
		Twitter I & \cite{you2015robust} & 1,269 & social & 5 per image& Sentiment & one sentiment category for each image\\
		Twitter II & \cite{borth2013large} & 603 & social & 3 per image & Sentiment & one sentiment category for each image\\
		VSO & \cite{borth2013large} & $\approx$500,000 & social & -- & Plutchik & one emotion category for each image\\
		MVSO & \cite{jou2015visual} & 7.36M & social & -- & Plutchik & one emotion category for each image\\
		Flickr I & \cite{yang2014your} & 354,192 & social & 6,735 & Ekman & one emotion category for each image\\
		Flickr II & \cite{katsurai2016image} & 60,745 & social & 3 per image& Sentiment & one sentiment category for each image\\
		Instagram & \cite{katsurai2016image} & 42,856 & social & 3 per image& Sentiment & one sentiment category for each image \\
		Emotion6 & \cite{peng2015mixed} & 1,980 & social & 432 & Ekman+neutral, VA & the discrete probability distribution\\
		FI & \cite{you2016building} & 23,308 & social & 225 & Mikels  & one DEC for each image\\
		IESN & \cite{zhao2016predicting} & 1,012,901 & social & 118,035 & Mikels, VAD & the emotion of involved users for each image \\
		T4SA& \cite{Vadicamo_2017_ICCVW} & 1,473,394 & social &- & Sentiment+neutral & one sentiment category for each image \\
		B-T4SA& \cite{Vadicamo_2017_ICCVW} & 470,586 & social &- & Sentiment+neutral & one sentiment category for each image\\
		Comics & \cite{she2019learning} & 11,821 & comic &10 (5f,5m) & Mikels & one DEC for each image\\
		Event & \cite{ahsan2017towards} & 8,748 & social &3 each image& Sentiment+neutral & one sentiment category for each image\\
		EMOTIC & \cite{kosti2017emotion} & 18,316 & social & 3 each image& Ekman, VAD  & one DEC and VAD values for each image\\
		EMOd & \cite{fan2018emotional} & 1,019 & natural & 3 & Sentiment+neutral  & object contour, object name, sentiment category\\
		WEBEmo & \cite{panda2018contemplating} & 268,000 & social & - & Parrott & one DEC for each image\\
		LUCFER & \cite{balouchian2019lucfer} & 3.6M & social & - & Plutchik, VAD, context & one DEC,  average VAD values, and context for each image \\
		FlickrLDL & \cite{yang2017learning} & 10,700 & social & 11 & Mikels  & the discrete probability distribution\\
		TwitterLDL & \cite{yang2017learning} & 10,045 & social & 8 & Mikels & the discrete probability distribution\\
		\bottomrule
	\end{tabular}
	\label{tab:Dataset}
\end{table*}

Another relevant concept worth mentioning is that emotion in response to multimedia can be expected, induced, or perceived emotion.
Expected emotion is the emotion that the multimedia creator intends to make people feel, perceived emotion is what people perceive as being expressed, while induced/felt emotion is the actual emotion that is felt by a viewer. We do not aim discussing the difference or correlation of various emotion models in this survey and believe that the achievements from psychology and cognitive science are beneficial for the AICA task.% The typical emotion models that have been widely used in AICA are summarized in Table~\ref{tab:EmotionModels}.

% %Bjoern: I miss the difference between perceived emotion and intended emotion, etc. In other words, could a column be added in the table what has been annotated? What viewers *personally* felt can be very different from what they think others will feel or an artist intended to make people feel. In music emotion recognition, this is quite an important distinction. So, for example: A scene could one make laugh, despite it being violent, so (s)he could label what he thinks is intended as emotion, what (s)he thinks most would feel, or what (s)he feels personally. Might be difficult to find as info in all databases, so instead, one could add a short comment...

%SZ: This is a good question. Actually, there is such a column in the IJCAI conference version. As you mentioned, we found that it is very difficult to find such info, as many papers do not contain the label details. We will dig into the datasets deeper if reviewers also ask so. Is that OK?

\section{Datasets}

\label{sec:Datasets}
In the early years, the affective datasets only contain small-scale images built from psychology or artistic communities.
%The affective datasets are built based on small scale images from psychology or artistic communities in the earlier years.
With the development of digital photography and online social networks, an increasing number of large-scale datasets have been created by crawling the images posted on Internet.
We summarize all the datasets for AICA in Table~\ref{tab:Dataset}.

%To further investigate different datasets, we conduct extensive experiments in Section~\ref{}.

\subsection{Brief Introduction to Different Datasets}
The International Affective Picture System (\textbf{IAPS})~\cite{lang1997international} is an image dataset for visual emotional stimuli used in experimental investigations of emotion and attention in psychology~\cite{lang1997international}.
The dataset contains 1,182 documentary-style natural color images with various contents or scenes, such as portraits, babies, animals, landscapes, \textit{etc.}
About one hundred college students took part in the VAD rating
%BS:
on
a 9-point scale.
The mean and standard deviation (STD) of scores for each image can be derived easily.
%Each image is associated with an empirically derived mean and standard deviation (STD) of VAD ratings in a 9-point rating scale by about one hundred college students.

The subset A of IAPS (\textbf{IAPSa})~\cite{mikels2005emotional} is collected from IAPS to characterize
%BS
the images by a
descriptive discrete emotion category.
Specifically, 203 negative images and 187 positive images are selected, and then labeled by twenty undergraduate participants.
To the best of our knowledge, it is the first affective image dataset which is labeled using
%BS:
a
discrete emotion category.

The \textbf{Abstract} dataset~\cite{machajdik2010affective} consists of 280 paintings which are combinations only of color and texture.
They are annotated by about 230 people, and each image is voted 14 times.
For each image, the category obtaining the most number of votes is regarded as the ground truth.
After filtering the images whose votes are inconclusive, 228 images are retained.

%The \textbf{Artphoto} dataset~\cite{machajdik2010affective} consists of 806 artistic photos collected from an art sharing site by considering names of emotional categories as keywords when searching.
The \textbf{Artphoto} dataset~\cite{machajdik2010affective} contains 806 artistic photos collected from an art sharing site. The photos are obtained by searching the site with the emotion categories as keywords.
The ground truth of each image is determined by the user who uploads it.

% The three small datasets including IAPSa, Abstract, and Artphoto are labeled according to eight emotion categories defined in Mikel's model~\cite{mikels2005emotional}.
% %
% In IAPSa, some images are labeled with over one emotion, while each image in Artphoto has only a single label.
% %
% All the 280 images in Abstract have discrete emotion distributions as labels.
% %
% However, only 228 images have
% %BS:
% a
% specific DEC.

The
%BS:
Geneva
affective picture database (\textbf{GAPED})~\cite{dan2011geneva} contains 730 pictures which are collected to make full use of visual emotion stimuli.
Several specific
%BS:
types of
negative or positive content are presented in these images.
The 520 negative images, 121 positive images, and 89 neutral images are labeled by 60 people ranging from 19 to 43 years (mean=24, STD=5.9).
%
%In addition, the continuous scales (\emph{valence}, \emph{arousal}, \emph{dominance}, \emph{etc.}) are rated from 0 to 100 points.
In addition, the continuous VA scales are rated from 0 to 100 points.

The \textbf{MART} dataset~\cite{alameda2016recognizing} contains 500 abstract paintings collected from more than 20,000 artworks of professional artists guided by an art historian.
There are 25 participants (11 females and 14 males) annotating these images with negative or positive
%BS:
rating.
Each person annotated 145 paintings on average.

The \textbf{devArt}~\cite{alameda2016recognizing} is a dataset of Amateur paintings from the deviantArt (dA)
website.
The 500 paintings created by 406 different authors are labeled by 60 people including 27 females and 33 males.

The \textbf{Twitter I} dataset~\cite{you2015robust} consists of 1,269 images.
%to valid the performance of trained model.
%
A total of 5 Amazon Mechanical Turk (AMT) workers were employed to label the images.
The dataset contains three subsets, including ``Five agree" (Twitter I-5), ``At least four agree" (Twitter I-4) and ``At least three agree" (Twitter I-3).
``Five agree" indicates that all the 5 AMT workers reached an agreement on the sentiment label of an image.
Twitter I-5 contains 882 images, while all the images obtain at least three same votes on sentiment.

The \textbf{Twitter II} dataset~\cite{borth2013large} includes 470 positive images and 133 negative images collected using over 20 twitter hashtags.
Three different labeling runs, namely image-based, text-based, and image-text based, were conducted by 3 random AMT workers (each worker for each run), respectively. The final selected images all receive unanimous sentiment votes.
%
%The final selected images all receive unanimous votes for sentiment label.

The images labeled with 1,553 ANPs in \textbf{VSO}~\cite{borth2013large} are retrieved and downloaded using the Flickr API.
The corresponding ANP should be contained in the title, tag, or caption of the image.
As psychological principles for construction of datasets, Plutchik's Wheel of Emotions covers 3 intensities based on 8 basic emotions.
\textbf{MVSO}~\cite{jou2015visual} is the extension of the VSO.
The dataset consists of more than 7.36M images annotated with ANPs from 12 different languages including Arabic, Chinese, Dutch, English, French, German, Italian, Persian, Polish, Russian, Spanish, and Turkish.
%w
A total of 4,342 English ANPs were constructed.
% to describe the images.
%BS:

\textbf{Flickr I}~\cite{yang2014your} is proposed to study the correlation between emotions and friends' discussions on the images.
It contains 354,192 images posted by 4,807 users, of which all the comments and tags are included.
To model the friends' interactions well, the detailed information of users is also recorded in the dataset, including
%BS:
ID, alias, and contact list.

\textbf{Flickr II}~\cite{katsurai2016image} and \textbf{Instagram}~\cite{katsurai2016image} are collected from
%BS:
Flickr and Instagram, respectively, based on query keywords.
The sentiment polarity labels are provided via online crowdsourcing.
Specifically, each image was shown to three random workers, who should choose a rating from the five scales including highly positive, positive, neutral, negative, and highly negative.
The final ground-truth of each image is determined by the major ratings of polarity.
After discarding the images that are labeled as `neutral' or received opposite sentiment annotations, 48,139 positive and 12,606 negative images are left in
%BS:
the
Flickr II dataset, while Instagram contains 33,076 positive and 9,780 negative images.

\textbf{Emotion6}~\cite{peng2015mixed} contains 1,980 images which are obtained from Flickr by six category keywords and corresponding synonyms.
Each image is annotated by 15 participants with both valence-arousal scores and discrete emotion distribution.
The categories include Ekman's six basic emotions~\cite{ekman1992argument} and neutral.

\textbf{FI}~\cite{you2016building} is a large-scale affective image dataset constructed based on Mikel's emotions.
All the images are collected from Flickr and Instagram with the eight emotions as search keywords.
A total millions of weakly labeled images are crawled.
After deleting noisy data, a total of 225 AMT workers were employed to assess the emotions of images.
Finally, 23,308 images receive at least three agreements from the assigned annotators.
The dataset is widely used in the field of AICA.
%A total of 225 AMT workers take part in

\begin{figure*}[t]
	\begin{center}
		\includegraphics[width=0.92\linewidth]{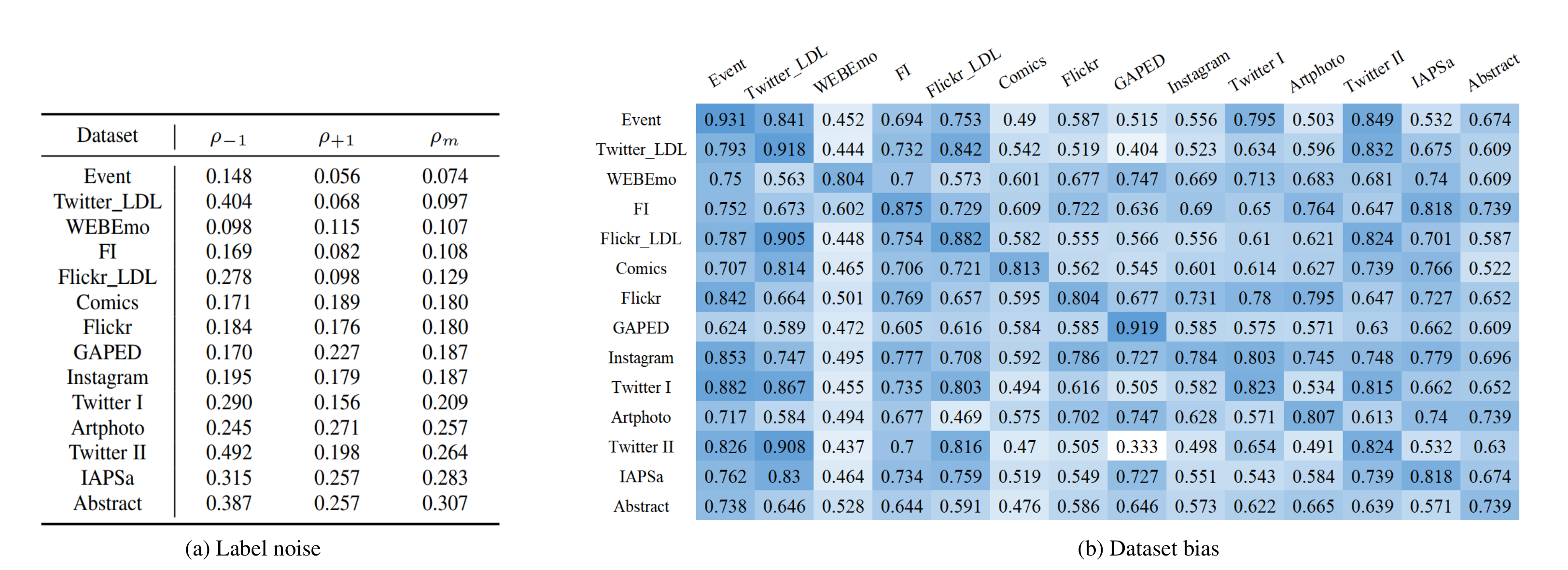}
		%\fbox{\rule{0pt}{2in} \rule{\linewidth}{0pt}}
		\vspace{-10pt}
		\caption{Quantitative comparison and ranking of different datasets. (a) Estimated label noise ratio of different datasets. $\rho_{-1}$ means the noise ratio of negative sentiment, and $\rho_{+1}$ means the noise ratio of positive sentiment. $\rho_m$ is the mean noise ratio. (b) The confusion matrix between different datasets on sentiment polarity classification, which can reflect the bias between any two datasets.}
		\vspace{-10pt}
		\label{fig:DatasetComparison}
	\end{center}
\end{figure*}

The \textbf{IESN} dataset~\cite{zhao2016predicting} consists of more than one million images crawled from Flickr uploaded by 11,347 users, and it is constructed to study the personalized emotion perception.
Therefore, various metadata of corresponding images are also collected, including tags, descriptions, comments, and uploaders' social context.
For each image, the labels of expected emotion from the uploader and actual emotion from the viewer are both generated.
In addition, by leveraging the VAD norms of 13,915 English lemmas~\cite{warriner2013norms}, the average
%BS:
values
of VAD are computed as label of DES.
According to the descriptions and the comments of
%BS:
the
images, the emotion distribution is also easy to obtain.

The \textbf{T4SA} dataset~\cite{Vadicamo_2017_ICCVW} consists of about one million tweets and corresponding images.
According to the textual sentiment classification, the images are classified into positive, negative, and neutral.
However, the dataset contains 501,037 positive, 214,462 negative, and 757,895 neutral images, which is imbalanced.
%
%The \textbf{B-T4SA} is a balanced subset of the T4SA, in which there are 156,862 images in each class.
%In order to obtain a balanced dataset,
As a balanced subset, \textbf{B-T4SA}~\cite{Vadicamo_2017_ICCVW} is extracted from T4SA, in which there
%BS
is an equivalent of
156,862 images in each class.

%BS:
The
\textbf{Comics} dataset~\cite{she2019learning} is composed of 11,821 images selected from seventy comics, including
%BS:
One Piece, Spiderman, Sponge Bob, The Avengers, \etc.
A total of 10 participants (mean age=20.3
%BS:
years)
were employed to label these images using Mikel's eight emotion categories.
The dataset is further divided into two subsets:
%BS:
A
Comics subset and
%BS:
a Manga subset.
%BS:
The
Comics subset includes samples from European and American
%BS:
comics
which are drawn in the realism style, while
%BS: too much to change to leave %BS everywhere...
the
Manga subset contains Asian comics with an abstract style.

The \textbf{Event} dataset~\cite{ahsan2017towards} has 8,748 images that are obtained from Microsoft Bing by using search keywords from 24 event categories.
The types of events are diverse, including personal and actual public events.
In the annotation process, an image could be retained if at least 2 out of 3 crowdworkers reach an agreement on its label.
The sentiment labels of these event images contain positive, negative, and neutral.

The \textbf{EMOTIC} dataset~\cite{kosti2017emotion} contains 18,316 images which are selected from the MSCOCO~\cite{lin2014microsoft} and Ade20k~\cite{zhou2019semantic} datasets and were downloaded via the Google search engine based on 26 emotional keywords.
The dataset has two types of annotated information.
One is the DES comprising 26 emotions, while the other is continuous VAD dimensions.
A total of 23,788 people (66\% males, 34\% females) are annotated in the images.

The \textbf{EMOd} dataset~\cite{fan2018emotional} consists of 1,019 emotional images, in which 321 images are selected from IAPS~\cite{lang1997international} and 698 images are collected via online image search engine.
Each image owns eye-tracking data collected from 16 participants, and is labeled with detailed information including object contour, object sentiment, semantic category, and high-level perceptual attributes such as image aesthetics and emotions.
Three undergraduate students were employed to label the characteristic of the objects in the images.
The emotion of the object is labeled by `neutral' when the agreements are different.

The \textbf{WEBEmo} dataset~\cite{panda2018contemplating} is a large-scale web emotion dataset constructed based on Parrott's hierarchical emotion model.
First, about 300,000 weakly labeled images are collected by searching with keywords for each of the 25 emotions.
Then, the duplicate images and those with non-English tags are removed, leading to 268,000 high-quality images.

The \textbf{LUCFER} dataset~\cite{balouchian2019lucfer} contains over 3.6M images, which are labeled with 24 emotional categories from Plutchik's model, 3 continuous emotional dimensions, and image contexts.
By combining contexts and emotions, a total of 275 emotion-context pairs are generated.
First, 80,649 images are collected from the \textit{wild}, and 35,239 images of them pass the AMT workers' validation.
With over 30 thousand images, a large-scale dataset is obtained employing \textit{Bing}’s feature available in its \textit{Image Search API}.

\textbf{Flickr\_LDL}~\cite{yang2017learning} and \textbf{Twitter\_LDL}~\cite{yang2017learning} are constructed to study the emotion ambiguity.
Each image is labeled by more than one viewer based on their own emotional reactions.
Flickr\_LDL contains 10,700 images extracted from VSO dataset.
A total of 11 participants were hired to view each image and label the images with one of the Mikel's eight emotions.
Twitter\_LDL is collected from Twitter with various sentiment keywords, and 8 viewers were employed to label these images within the same eight emotions.
Finally, 10,045 images are retained after deduplication.

%
% \begin{figure*}[t]
% 	\begin{center}
% 		\includegraphics[width=0.7\linewidth]{dataset_bias_2.pdf}
% 		%\fbox{\rule{0pt}{2in} \rule{\linewidth}{0pt}}
% 		\caption{The confusion matrix between different datasets on sentiment polarity classification, which can reflect the bias between any two datasets.}
% 		\label{fig:dataset_bias_2}
% 	\end{center}
% \end{figure*}

\subsection{Comparison Among Different Datasets}

%The aforementioned datasets have different characteristics.
%
%With the development of AICA, an increasing number of datasets with large scale and more categories are constructed.
%
%In the early years, the affective datasets only contain hundreds of images. With the development of AICA, an increasing number of datasets with large scale and more categories are constructed.
Here we compare several released datasets from the perspectives of label noise and dataset bias for readers to better understand how to select required datasets in real applications.

\begin{table*}[!t]
	\centering\scriptsize
	\caption{Summary of hand-crafted features on different levels that have been used for AICA. `\# Feat' indicates the dimension of each feature.}
	\begin{tabular}
		{lll p{12cm} r}
		\toprule
		\textbf{Feature} & \textbf{Ref} & \textbf{Level} & \multicolumn{1}{l}{\textbf{Short description}} & \textbf{\# Feat} \\
		\hline
		WLDLV & \cite{wei2004image} & low  & orientation and length information of lines & 12\\
		EFS & \cite{wei2006image} & low & luminance-warm-cool fuzzy histogram, saturation-warm-cool fuzzy histogram, luminance contrast & 10, 7, 2\\
		Eleven Groups & \cite{zhang2011analyzing} & low & shape, edge, texture, polynomial, image statistics & 691\\
		LOW\_C & \cite{patterson2012sun} & low  & Gist, HOG2x2, self-similarity and geometric context color histogram features & 17,032\\
		Elements & \cite{machajdik2010affective} & low  & color: mean saturation, brightness and hue, emotional coordinates, colorfulness, color names, Itten contrast, Wang's semantic descriptions of colors, area statistics; texture: Tamura, Wavelet and gray-level co-occurrence matrix & 97\\
		MPEG-7 & \cite{lee2011fuzzy} & low  & color: layout, structure, scalable color, dominant color; texture: edge histogram, texture browsing &  $\approx$200 \\
		Shape & \cite{lu2012shape} & low & line segments, continuous lines, angles, curves &  219\\
		IttenColor & \cite{sartori2015s} & low & color co-occurrence features and patch-based color-combination features & 16,485\\
		%\hline
		Attributes & \cite{patterson2012sun} & mid & scene attributes & 102 \\
		Sentributes & \cite{yuan2013sentribute} & mid & scene attributes, eigenfaces & 109 \\
		Constructs & \cite{lu2017investigation} & mid & roundness, angularity, complexity  & 3 \\
		Composition & \cite{machajdik2010affective} & mid & level of detail, low depth of field, dynamics, rule of thirds & 45 \\
		Aesthetics & \cite{wang2013interpretable} & mid & figure-ground relationship, color pattern, shape, composition & 13 \\
		Principles & \cite{zhao2014exploring} & mid & principles-of-art: balance, contrast, harmony, variety, gradation, movement &  165\\
		SIFT & \cite{rao2016multi} & mid & bag-of-visual-words on SIFT, latent topics & 330 \\
		%\hline
		FS & \cite{machajdik2010affective} & high & number of faces and skin pixels, size of the biggest face, amount of skin w.r.t. the size of faces & 4 \\
		ANP & \cite{borth2013large} & high & semantic concepts based on adjective noun pairs & 1,200 \\
		Expressions & \cite{yang2010exploring} & high & automatically assessed facial expressions (anger, contempt, disgust, fear, happiness, sadness, surprise, neutral) & 8 \\
		HLCs & \cite{ali2017high} & high & object information and scene information & 1,205 \\
		\bottomrule
	\end{tabular}
	\label{tab:HandCraftedFeatures}
\end{table*}

\subsubsection{Label Noise}

For quantitative comparison in terms of label noise, we estimate the noise rate by pre-training CNN with softmax loss~\cite{liu2015classification}.
It is assumed that the probability of positive (+1) images being assigned to negative (-1) is $\rho_{+1} = p(\hat{y}=-1|y=+1)$, where $y$ represent the ground truth and $\hat{y}$ is the predicted label.
Similarly, the probability of negative images being assigned to positive is $\rho_{-1} = p(\hat{y}=+1|y=-1)$.
If there is no label noise, the values of $\rho_{-1}$ and $\rho_{+1}$ will approach 0.
%
%Following [20], [22], $\rho_{+1}$ and $\rho_{-1}$ can be computed as follows:
%$\rho_{+1} = 1 - \max p(\hat{y}_i = +1 | x_i )$, $\rho_{-1} = 1 - \max p(\hat{y}_i = +1 | x_i )$, which represent the upper bounded of noise.
%
In Fig.~\ref{fig:DatasetComparison} (a), we estimate the values of $\rho_{+1}$ and $\rho_{-1}$ using the algorithm in~\cite{wang2018visual, liu2015classification}.
First, we resort images based on the predicted probability values from small to large:
\begin{equation}
\begin{aligned}
\hat{p}(\hat{y}|x_{n+1})\geq\hat{p}(\hat{y}|x_n), \hat{p}(\hat{y}|x_{N_{\hat{y}}}) = \max\hat{p}(\hat{y}|x_n),
\end{aligned}
\end{equation}
where $n = 1,2,\cdots,N_{\hat{y}}$, and $N_{\hat{y}}$ denotes the number of images that is predicted as $\hat{y}$.
Second, we build a fitting Gaussian function $g(\cdot)$ between the predicted probability and the corresponding numbers of it:
\begin{equation}
n_{\hat{p}(\hat{y}_n|x_n)}= g_{\hat{y}_n}(\hat{p}(\hat{y}_n|x_n)), N_{\hat{p}(\hat{y}_n|x_n)}=\int_0^1 g_{\hat{y}_n} (\hat{p}(\hat{y}_n|x_n)),
\end{equation}
where $n_{\hat{p}(\hat{y}_n|x_n)}$ is the number of images that are predicted as $\hat{y}_n$ with probability $\hat{p}(\hat{y}_n|x_n)$.
Third, the noise ratio is regarded as the deviation between the probability that obtains the maximum value on the fitting function and 1:
\begin{equation}
\begin{aligned}
\rho_{+1}= 1-\arg\max g_{+1}(\hat{p}(\hat{y}_n=+1|x_n)),\\
\rho_{-1}= 1-\arg\max g_{-1}(\hat{p}(\hat{y}_n=-1|x_n)),
\end{aligned}
\end{equation}
Finally, we compute the mean noise ratio $\rho_m$ weighted by the proportion of each category:
\begin{equation}
\rho_m=\sum\nolimits_i^c v_i\cdot \rho_{i},
\end{equation}
where $v_i$ is the proportion of images from the $i$-th category, $\rho_{i}$ is the noise ratio of the $i$-th category, and $c$ is the number of categories.
From the table, we can observe that Event is the dataset with the least label noise.
However, Abstract has the largest label noise.
It is mainly because the abstract images are difficult to distinctly understand, leading to more label noise.
%
%

% \begin{table}[t]
% 	\centering\scriptsize
% 	\setlength\tabcolsep{1.5pt}
% 	\caption{Estimated label noise ratio of different datasets. $\rho_{-1}$ means the noise ratio of negative sentiment, and $\rho_{+1}$ means the noise ratio of positive sentiment. $\rho_m$ is the mean noise ratio.}
% 	\vspace{-5mm}
% 	\begin{center}
% 		\begin{tabular}{ p{1.5cm}<{\centering} |p{1.2cm}<{\centering} p{1.2cm}<{\centering}  p{1.2cm}<{\centering}}
% 			\toprule
			
% 			Dataset	     &  $\rho_{-1}$&$\rho_{+1}$   & $\rho_m$          \\ \midrule
% 			FI           & 0.169   & 0.082    &  0.108       \\
% 			Flickr\_LDL  & 0.278  & 0.098     &  0.129       \\
% 			Twitter\_LDL & 0.404  & 0.068     &  0.097       \\
% 			IAPSa        & 0.315  & 0.257     &  0.283       \\
% 			Abstract     & 0.387  & 0.257     &  0.307       \\
% 			Artphoto     & 0.245  & 0.271     &  0.257       \\
% 			Comics       & 0.171  & 0.189     &  0.180       \\
% 			GAPED        & 0.170  & 0.227     &  0.187       \\
% 			Event        & 0.148  & 0.056     &  0.074       \\
% 			Twitter I    & 0.290  & 0.156     &  0.209       \\
% 			Twitter II   & 0.492  & 0.198     &  0.264       \\
% 			Flickr       & 0.184  & 0.176     &  0.180       \\
% 			Instagram    & 0.195  & 0.179     &  0.187       \\
% 			WEBEmo       & 0.098  & 0.115     &  0.107       \\
% 			\bottomrule
% 		\end{tabular}
% 	\end{center}
% 	\label{dis_SCU}
% \end{table}

\subsubsection{Dataset Bias}
To explore the dataset bias of emotion recognition, we conduct extensive transfer learning experiments among various datasets by training a classifier on one dataset and testing on another.
These datasets have wide ranges on the number of images, ranging from a few hundred to a few hundred thousand.
Meanwhile, these datasets include various types, such as natural images and abstract images.
We split each dataset into 80\% training images and 20\% test images.
For a fair experiment, we fine-tune a ResNet-50 on each dataset and test the model on all the datasets.

The confusion matrix is shown in Fig.~\ref{fig:DatasetComparison} (b), where each row represents the results of one model on different datasets. We have some observations as follows.
First, there is a larger bias between datasets that belong to different types, such as abstract, comic, and natural images.
For example, the types of Abstract, Comics, and GAPED are abstract, comic, and natural, respectively.
The model trained on Comics only obtains 0.522 and 0.546 accuracy on Abstract and GAPED datasets.
Second, the bias between the two datasets is not mutual.
This means that the evaluated bias is smaller when the cleaner one of two datasets is regarded as the target.
For instance, as the shown experimental results between the Event (the cleanest dataset, $\rho_m$ = 0.074) and other datasets, the results are better when Event is the target dataset.
Third, the similarity of class distribution can influence the dataset bias.
For example, Twitter II and Flickr\_LDL (Twitter\_LDL) have similar class distribution, where there are significantly more positive images than negative images.
However, Flickr and Instagram have the balanced class distribution, which is different from that of Twitter II.
Therefore, it is observed that the model trained on Twitter II has better performance on Flickr\_LDL and Twitter\_LDL than that on Flickr and Instagram.
Finally, as the most widely-used dataset, FI has the best generalization ability among these datasets.
As shown in the first row of Fig.~\ref{fig:DatasetComparison} (b), the model trained on FI dataset obtains over 60\% classification accuracy on all the datasets.

\section{Emotion Feature Extraction}
\label{sec:Features}
To describe image emotion with informative representations, many studies explore extraction of various types of features.
In terms of hand-crafted features, we introduce the designed features on different levels.
%based on psychology theories.
%
Besides, we review the emerging deep features in recent years with the development of CNNs.

\subsection{Hand-crafted Features}
The hand-crafted features on different levels focus on different aspects, as summarized in Table~\ref{tab:HandCraftedFeatures}.
%

%We summarize the hand-crafted features for visual emotion in Table~\ref{tab:HandCraftedFeatures}.
%
%The features of different levels focus on different aspects, and their specific performances on different tasks are shown in the Tables~\ref{tab:Classification} and ~\ref{tab:Distribution}.
%
%Corresponding analysis for results is provided in Section~\ref{sec:Methods}.

\subsubsection{Low-level Features}
Various low-level features were designed to represent emotional content in the early years, although they lack reasonable interpretation.
As one pioneering study on hand-crafted features,~\cite{wei2004image} explores the relationship between line directions and image emotion.
Specifically, horizontal lines relate to the static horizon and always express the relaxation and calmness, while the direct and clear vertical lines convey eternality and dignity.
Lines with various directions can convey different emotions.
With the degree of long, thick, and straight lines increasing, the expressed emotions will be stronger.
However, to capture more informative information, guided by the theories of color psychology, \citeauthor{wei2006image}~\cite{wei2006image} constructed three kinds of image features in an orthogonal three-dimensional emotion factor space, respectively.
The features include luminance-warm-cool representation, saturation-warm-cool-contrast representation, and contrast-sharpness representation.
By mining these proposed emotional information, they designed an emotion-based image retrieval system.
%
%According to the local statistics information of images, textural features such as Wiccest~\cite{geusebroek2006compact} and Gabor~\cite{bovik1990multichannel} are extracted for each sub-region of an image to train a support vector machine (SVM) for each category.

When it comes to hand-crafted features, one cannot ignore a milestone~\cite{machajdik2010affective}, in which different types of features are combined.
Particularly, color and texture are the representative low-level features in composition.
%Inspired by the theories of psychology and art, Machajdik and Hanbury~\cite{machajdik2010affective} designed combined features, including color and texture, for image emotion classification.
%
Color is represented with a 70-dimensional vector consisting of eight kinds of statistical information, while texture is encoded with a 27-dimensional vector containing three types of image statistics.
Later in~\cite{lee2011fuzzy}, another complex feature combination consisting of color and texture, named MPEG-7, is proposed.
Besides, a fuzzy similarity relation is applied to computing the weights of different components.
%MPEG-7 descriptors, which are complex high-dimensional vectors, are used as features to recognize the emotion of images.
%
%Four color-related ideas and two texture-related ideas are included in MPEG-7.
%
\citeauthor{zhang2011analyzing}~\cite{zhang2011analyzing} listed eleven groups of features referring to texture, shape, and further, which are extracted from several transforms of the image.
The influence of the visual shapes on the image emotions is systematically explored in~\cite{lu2012shape}.
The experimental results demonstrate the effectiveness of shape features for emotion prediction.
%Besides, they constructed the concepts of roundness-angularity and simplicity-complexity as the important elements of distinguishing emotions.
%
%BS: Please spell out once GIST, HOG
%SZ: spell out HOG here
In addition, Gist, 2$\times$2 Histogram of Oriented Gradients (HOG), self-similarity, and geometric context color histogram features are widely used due to their ability to represent the distinct visual scenes~\cite{patterson2012sun}.
Based on Itten's color wheel, \citeauthor{sartori2015s}~\cite{sartori2015s} investigated the different color combinations in abstract paintings, and used the factors to analyze the emotions evoked in the viewers.

\subsubsection{Mid-level Features}
Compared with low-level features, mid-level ones are easier to understand by humans, and they can largely bridge the gap between low-level features and high-level emotions.
\citeauthor{patterson2012sun}~\cite{patterson2012sun} designed a large-scale attribute database, named the SUN attribute database, which consists of 102 attributes that belong to different types, including materials, surface properties, and others.
Based on these the mid-level attributes, \citeauthor{yuan2013sentribute}~\cite{yuan2013sentribute} proposed an image emotion recognition algorithm, named Sentribute, in which the facial expression detected based eigenface is also added as a crucial element to determine the polarity of emotions.

As an essential feature in artworks, harmonious composition~\cite{machajdik2010affective} is introduced for emotion representation.
Guided by the theories of art, \citeauthor{wang2013interpretable} ~\cite{wang2013interpretable} designed more interpretable and understandable features, in which the description of the contrast between the figure and the ground is included.
Apart from the studies that extract features of the overall image with the single scale, there are also some researchers that pay attention to mining informative representation in the multi-scale blocks of each image.
For instance, \citeauthor{rao2016multi}~\cite{rao2016multi} used two different image segmentation
%BS:
types
for extracting multi-scale blocks in each image.
With SIFT as the basic feature, they employed bag-of-visual-words (BoVW) to encode each block, and then adopted probabilistic latent semantic analysis to further estimate the latent topic used as a mid-level representation.
This study reveals that the BoVW can well model the emotional information of different local regions, and the features directly extracted from the whole image may lead to wrong classification results.
Based on different artistic principles, the combined emotion representation named principles-of-art is proposed~\cite{zhao2014exploring}, consisting of balance, emphasis, harmony, variety, gradation, and movement.
As the milestone of mid-level representation, principles-of-art features obtained the state-of-the-art performance at that time.
Without features of hundreds of dimensions, three visual characteristics, including roundness, angularity, and visual complexity, are proposed in~\cite{lu2017investigation}, each of which is only a
%BS:
one-dimensional
scalar.
It has been proved that these mid-level representations are effective when used to recognize image emotions.
%For the three mid-level representations, it has been demonstrated that they can be used to recognize emotions well.

\subsubsection{High-level Features}
High-level features refer to the semantic information of images, which are easy to understand and can directly evoke emotions in viewers.
In~\cite{machajdik2010affective}, \citeauthor{machajdik2010affective} extracted content that draws the attention of the viewers and has great effects on the emotions, including human faces and skin.
Facial expression, an important high-level feature, always plays a decisive role in the process of evoking emotions.
It is usually classified
%BS:
following Ekman's `Big Six'
into six basic emotions, which include anger, disgust, fear, happiness, sadness, and surprise.
In~\cite{yang2010exploring}, based on the compositional features of image patches, \citeauthor{yang2010exploring} detected and analyzed the categories of facial expression.
%
%The facial expressions can be classified into
%

As a milestone, a large visual sentiment ontology named SentiBank is proposed by \citeauthor{borth2013large}~\cite{borth2013large}.
It contains 1,200 concepts, and each concept represents an ANP, like \textit{beautiful flower}, which provides powerful semantic representation.
First, based on 24 emotions defined in Plutchik's theory, the authors retrieved related adjectives and paired them with frequently used nouns.
The final remaining 1,200 ANPs cover 178 adjectives and 313 nouns after filtering the redundant ANPs.
ANPs provided a novel solution to bridge the ``affective gap'', because they are easy to be mapped into emotions.
Later,
%BS:
as
the extension of the SentiBank, a large-scale multilingual visual sentiment ontology (MVSO) is proposed in~\cite{jou2015visual}.
Particularly, there are 4,342 English ANPs in MVSO.
In~\cite{ali2017high}, high-level concepts (HLCs), including objects and places, are introduced to bridge the affective gap between image content and evoked emotion.
The HLCs are explicitly derived from pre-trained CNNs, and subsequently, a linear admixture model is employed to capture the relations between emotions and HLCs.
%
%Finally, support vector regression (SVR) is used to obtain flexible emotional responses in viewers.

%and list the representative studies based on hand-crafted features and the corresponding experimental results on different datasets in the first part of Table~\ref{tab:HandCraftedMethods}.
%
%It is observed that high-level features show good performance when used for different tasks, including classification and retrieval.
%
%For the abstract paintings, low-level features (such as Elements and IttenColor) that focus on color, texture, \etc perform well, while mid-level features (such as Principles) are effective for artistic photos.
%

\subsection{Deep Features}
In recent years, with the rapid development of CNNs, learning-based features have shown superior performance in the
%BS:
field
of AICA.
In the beginning, each region in an image is treated equally in the learning process, and global features are extracted for different tasks.
%
%Later, based on the theories of psychology, emotional content can draw more human attention.
%
Later, based on the theories of psychology that emotional content is always involved in some important regions, more and more studies have focused on how to extract informative local features.
%
%Therefore, more and more studies focus on how to extract informative local features.

\subsubsection{Global Features}
Based on a deep CNN model, the classifiers of the 1,200 ANP concepts are trained using Caffe.
The newly trained deep model named DeepSentiBank~\cite{chen2014deepsentibank} performs better than non-deep SentiBank on sentiment prediction.
Benefiting from transfer learning, \citeauthor{xu5731visual}~\cite{xu5731visual} transferred the parameters of a CNN trained on the large-scale dataset (ImageNet) to the task of predicting sentiments.
They extracted a 4096-dimensional representation from the fully connected (FC) layer FC7 and a 1000-dimensional representation from FC8, respectively, as the image-level features.
The experimental results on Twitter II dataset demonstrate that the features from FC7 exhibit an advantage in describing the emotional content, because the activations from the FC7 layer can characterize more aspects of the image than those of FC8 layer.

The progressive CNN (PCNN)~\cite{you2015robust} is another milestone, which is pre-trained using half-a-million weakly labeled images from SentiBank.
In the learning process, the training instances that have large difference between the two polarities are kept for the next round of training.
With the iterative training, the noisy data can be removed progressively, such that the trained model is more robust when transferring to those small-scale strongly labeled datasets.
In~\cite{chen2015learning}, \citeauthor{chen2015learning} explored two methods of obtaining image features.
One method uses off-the-shelf CNN (pre-trained on ImageNet) features to train a one-vs-all linear SVM classifier.
The other method is to initialize the parameters of the pre-trained AlexNet and replace the 1,000-way classification layer with 8-way emotion category outputs.
Subsequently, the network can be trained end-to-end from raw images to the specific classes.

High-level semantic features may be not enough as emotional representation in some images, especially in abstract paintings.
To capture different types of information in the images, some studies integrate the multi-level deep features generated in CNNs.
\citeauthor{rao2016learning}~\cite{rao2016learning} proposed an end-to-end architecture that consists of three parallel neural networks, including an AlexNet, an aesthetics CNN (A-CNN), and a texture CNN (T-CNN).
%
%The A-CNN contains four convolutional layers, while T-CNN includes two convolutional layers.
%
Before being fed into the network, the images are segmented into different levels of patches.
Subsequently, the three sub-networks are exploited to extract deep representations at three-levels, respectively, \ie, image semantics, image aesthetics, and low-level
visual features.
\citeauthor{zhu2017dependency}~\cite{zhu2017dependency} extracted the multi-level features from different layers in CNNs.
The output from each layer is fed into a bidirectional gated recurrent unit
 (Bi-GRU) model to exploit the dependency between them.
Finally, the features output from the two ends are concatenated as the emotional representations.
It is further considered that a Gram matrix can capture powerful texture features~\cite{gatys2015texture}; hence,  \citeauthor{yang2018retrieving}~\cite{yang2018retrieving} proposed a sentiment representation consisting of elements in Gram matrices from different layers.

\subsubsection{Local Features}
To emphasize the informative regions that contain attractive emotional content, local features are drawing more and more attention in recent studies.
Considering fine-grained details, the features of local patches are extracted at multiple scales in~\cite{chen2015learning}.
Next, they are aggregated with the Fisher Vector for a more compact representation.
In~\cite{liu2016improving}, apart from investigating general emotion content using multiple instance learning, \citeauthor{liu2016improving} also detected facial expression and emotional objects to constitute the emotion factors when computing visual saliency.

How to find the crucial emotion-related regions based on an image-level label is a question worth exploring.
Based on the descriptive visual attributes, \citeauthor{you2017visual}~\cite{you2017visual} adopted an attention model to discover the local regions that evoke sentiment in viewers, and then extracted the features of them to improve the performance on visual sentiment analysis.
\citeauthor{yang2018visual}~\cite{yang2018visual} utilized an off-the-shelf object detection tool to generate bounding box candidates.
After removing the redundant proposals, the selected regions have a high probability of containing an object and accordingly obtain a high sentiment score.
Further, the features of selected regions and the holistic images are jointly used for classification.
However, the process of selecting proper image regions is time-consuming in this work.
To simplify and improve the step of selecting informative regions, a unified CNN that contains a classification branch and a detection branch is proposed in~\cite{yang2018weakly}.
In the detection branch, the soft sentiment map is generated by combining all the class-wise feature responses.
The comprehensive localized information can be derived by coupling the holistic feature and the sentiment map.
Later in~\cite{zhao2019pdanet}, both spatial and channel-wise attended features are incorporated into the final representation for visual emotion regression in VAD space.
To effectively utilize various information from multiple layers, \citeauthor{rao2019multi}~\cite{rao2019multi} proposed a multi-level region-based CNN framework to find the emotional response of the local regions.
First, the feature pyramid network (FPN) is employed to extract multi-level deep representations.
Following this, the regions of interest (ROIs) are detected based on the region proposal method, and their features in multiple levels are concatenated for image emotion classification.
The work has achieved the best classification performance on several benchmark datasets up to now.
To obtain an informative feature embedding for affective image retrieval, \citeauthor{yao2019attention}~\cite{yao2019attention} conducted polarity- and emotion-specific attention on the lower layers and higher layers, respectively.
The attended features from different layers are integrated by cross-level bilinear
%BS:
pooling
to generate the final representation.

\begin{table}[!t]
	\centering\scriptsize
	\caption{Experimental comparison between local and global features measured by average classification accuracy and rank. `L' denotes the local features, and `G' denotes the global features. FI2 denotes the binary sentiment classification results on the FI dataset, and FI8 denotes the classification results of eight emotions on FI (the same below).}
	\begin{center}
	\resizebox{\linewidth}{!}{%
		\begin{tabular}{ c |c c c | c c c}
		%\begin{tabular}{ p{1.8cm}<{\centering} |p{1.5cm}<{\centering} p{1.5cm}<{\centering}  p{1.5cm}<{\centering} | p{1.5cm}<{\centering} p{1.5cm}<{\centering} p{1.5cm}<{\centering}}
			\toprule
			\multirow{2}{1cm}{\textbf{Dataset}} &  \multicolumn{3}{c|}{\textbf{WSCNet}~\cite{yang2018weakly}}      & \multicolumn{3}{c}{\textbf{PDANet}~\cite{zhao2019pdanet}}   \\  \cline{2-7}
			             &  L    & G   & L+G  &   L    & G     & L+G            \\ \hline
			FI2          & 0.894 (3) & 0.894 (3) & 0.896 (1) &  0.807 (3)  & 0.876 (2) & 0.878 (1)           \\
			FI8          & 0.671 (3)  & 0.675 (2) & 0.679 (1) & 0.606 (3)  & 0.696 (1) & 0.694 (2)           \\
			Flickr\_LDL  & 0.697 (3) & 0.707 (2) & 0.709 (1)  & 0.592 (3) & 0.703 (1) & 0.703 (1)           \\
			Twitter\_LDL & 0.764 (3) & 0.773 (1) & 0.766 (2) &  0.725 (3) & 0.762 (2) & 0.763 (1)           \\
			Comics       &  0.531 (3) &  0.532 (2) & 0.542 (1) &  0.263 (3) & 0.595 (1)  & 0.588 (2)  \\
			GAPED        & 0.899 (2)  & 0.889 (3)  & 0.919 (1)  &  0.697 (3) &  0.939 (2)  & 0.950 (1)   \\
			Event        & 0.938 (2) & 0.937 (3) & 0.948 (1) &  0.791 (3) &  0.937 (2) & 0.946 (1)  \\
			Flickr       & 0.800 (3)  &   0.801 (2) &  0.807 (1) &  0.757 (3) & 0.808 (2)   &  0.819 (1)    \\
			Instagram    & 0.804 (3) & 0.816 (1) & 0.804 (3)  &  0.672 (3)  &  0.811 (1) & 0.807 (2)  \\
			Twitter I    & 0.819 (3) & 0.827 (1) & 0.827 (1) &  0.606 (3) &  0.839 (2) & 0.858 (1)  \\
			Twitter II   & 0.824 (1) & 0.824 (1) & 0.815 (3)  &  0.815 (1) &  0.815 (1)  & 0.815 (1)  \\ \hline
			Average rank  &      2.636 (3)     &    1.909 (2)       &  1.455 (1)     &  2.818 (3)     &   1.545 (2)  & 1.273 (1)\\
			\bottomrule
		\end{tabular}
		}
	\end{center}
	\label{local_global}
\end{table}

\begin{table*}[!t]
	\centering\scriptsize
	\setlength\tabcolsep{1.5pt}
	\caption{Experimental results of different features on widely-used datasets. For each feature, the average results of different classifiers are also reported.}
	\vspace{-5mm}
	\begin{center}
		\begin{tabular}{ p{1.5cm}<{\centering} | p{0.71cm}<{\centering} p{0.71cm}<{\centering} p{0.71cm}<{\centering} p{0.71cm}<{\centering}| p{0.71cm}<{\centering} p{0.71cm}<{\centering} p{0.71cm}<{\centering} p{0.71cm}<{\centering} |p{0.71cm}<{\centering}p{0.9cm}<{\centering} p{0.71cm}<{\centering} p{0.71cm}<{\centering} | p{0.71cm}<{\centering}p{0.71cm}<{\centering} p{0.71cm}<{\centering} p{0.71cm}<{\centering}| p{0.71cm}<{\centering} p{0.71cm}<{\centering} p{0.71cm}<{\centering} p{0.71cm}<{\centering}}
			\toprule
			\multirow{2}{1cm}{\textbf{Dataset}}     &  \multicolumn{4}{c|}{\textbf{PAEF}~\cite{zhao2014exploring}}      & \multicolumn{4}{c|}{\textbf{Sun attribute}~\cite{patterson2012sun}}       & \multicolumn{4}{c|}{\textbf{SentiBank}~\cite{borth2013large}} &       \multicolumn{4}{c|}{\textbf{MVSO}~\cite{jou2015visual}} &\multicolumn{4}{c}{\textbf{P-VGG}~\cite{simonyan2015very}}  \\  \cline{2-21}
			& $k$NN   &        NB       &    SVM   &  Avg    & $k$NN   &        NB       &    SVM & Avg    & $k$NN   &     NB    &    SVM & Avg  & $k$NN  &  NB    & SVM& Avg   &  $k$NN  &  NB    & SVM &Avg  \\ \hline
			%Abstract   & 0.244 & 0.278 & 0.256  & 0.261 & 0.257 & 0.323 &   0.158 & 0.175 & 0.279 &   0.222  & 0.283     &   0.317 & 0.265 & 0.278 & 0.326 \\
			%Artphoto     &0.213  & 0.248 & 0.247 & 0.213 & 0.239  & 0.285  & 0.194  & 0.120 & 0.242 & 0.254 & 0.206  & 0.322 & 0.234 & 0.308 &0.312  \\
			%IAPSa     & 0.142&0.196 & 0.191&   0.124  & 0.183 & 0.194  & 0.152 & 0.155 & 0.233 & 0.186  & 0.194  & 0.179 & 0.198 &0.280  & 0.275 \\
			Emotion6     & 0.246 & 0.288 & 0.359 & 0.298 & 0.268 & 0.323  & 0.306 & 0.299 & 0.283  & 0.290 & 0.342 & 0.305  & 0.431  & 0.460 & 0.508   & 0.466   & 0.429 & 0.453 & 0.510 & 0.464 \\
			FI2         & 0.687 & 0.733 & 0.730 & 0.717 & 0.698 & 0.697  & 0.739 & 0.711 &   0.603 & 0.815  & 0.815  &  0.744  & 0.797  & 0.706   & 0.831   & 0.778   & 0.820 & 0.737 & 0.851 & 0.803 \\
			FI8         & 0.286 & 0.299 & 0.343 & 0.309 & 0.300 & 0.271  & 0.372 & 0.314 &   0.445 & 0.288 & 0.506 & 0.413     & 0.529  & 0.389   & 0.600   & 0.506  & 0.556 & 0.497 & 0.630 & 0.561 \\
			Flickr       & 0.627 & 0.640 & 0.674 & 0.647  &   0.634  & 0.639 & 0.683 &  0.652   &  0.581 & 0.608  & 0.694 & 0.628 & 0.697 & 0.699    & 0.771    & 0.722   & 0.707 & 0.699 & 0.777 & 0.728  \\
			Instagram       & 0.556  & 0.589 & 0.638 & 0.594  & 0.561 & 0.586  & 0.631 &  0.593 &   0.584  & 0.576 & 0.662 &  0.607 & 0.667  & 0.717    & 0.750    & 0.711   & 0.701  & 0.712  & 0.772 &  0.728  \\
			Twitter I       & 0.593 & 0.633 & 0.675 &  0.634 & 0.565 & 0.615  & 0.643 & 0.608   &  0.526 & 0.564   & 0.602 & 0.564 & 0.696   &  0.606   & 0.775    & 0.692   & 0.674  & 0.729  & 0.741 & 0.715 \\
			Twitter II       &  0.659 & 0.777 & 0.777 &  0.738 & 0.672 & 0.606  & 0.777 & 0.685  &  0.632 & 0.661 & 0.777 & 0.690 &  0.651  &  0.777   & 0.777    & 0.735   & 0.631 & 0.643  & 0.792 & 0.689 \\
			\bottomrule
		\end{tabular}
		\vspace{-10pt}
	\end{center}
	\label{nonlearning}
\end{table*}

\subsubsection{Comparison Between Local and Global Features}
To fairly evaluate the effectiveness of local features and global features, we conduct the comparison experiments in Table~\ref{local_global} based on WSCNet and PDANet, which are the state-of-the-art methods that consider local regions by combining both local features and global features into the final representation.
We conduct the comparison experiments for local, global, and combined local-global features for the two representative methods.
In the last row of the table, we provide the results of average rank, which demonstrate that the global features outperform the local features in general.
Besides, for most datasets, the results using both local and global features are better than that using only one type of features.
It is mainly because both local and global features can determine the emotions to some extent, and some local regions may generate emotional prioritization effect rather than sole effect.
Therefore, local features should be effectively integrated with global features for the more discriminative representations~\cite{compton2003interface}.

\subsection{Quantitative Feature Comparison}
In Table~\ref{nonlearning}, we evaluate the performance of different features based on different classifiers.
The results of six widely-used datasets are reported.
Note that FI is regarded as a dataset that simultaneously has two sentiment categories and eight emotion categories.
The hand-crafted features include PAEF~\cite{zhao2014exploring}, Sun attribute~\cite{patterson2012sun}, and SentiBank~\cite{borth2013large}, while off-the-shelf deep features are extracted from MVSO~\cite{jou2015visual} and pre-trained VGGNet-16~\cite{simonyan2015very}.
Each type of feature is used to train three classifiers, including $k$NN, Naive Bayes (NB), and support vector machine (SVM).
In Sun attribute, we extract the four types of features, including GIST, HOG 2$\times$2, self-similarity, and geometric context color histogram features.
%
%Finally, we reduce the features into 256 dimension
%
In pre-trained VGGNet-16, we extract 4096-dimensional features from the last layer.
Note that the features of Sun attribute and pre-trained VGGNet-16 are both reduced to 256-dimension.
We report the average results of different classifiers for the same feature to fairly investigate the representation ability of each feature.
From the results of the same classifiers and the average results of different classifiers, it is observed that deep features obtain the best performance, which is also demonstrated in traditional computer vision tasks, such as image classification and object detection.
Besides, high-level features (\eg SentiBank) perform better than middle-level features (\eg PAEF) in most cases.
It is mainly because high-level features are more related to the emotional semantics.
For example, SentiBank is constructed based on ANPs, where adjective can be better mapped into sentiment.

\section{Learning Methods for Different Tasks}%{Task}
\label{sec:Methods}

In this section, we review the learning methods of recent two decades on AICA, in which significant development has been obtained on different AICA tasks, including dominant emotion recognition, personalized emotion prediction, emotion distribution learning, and learning from noisy data or few labels.
%The corresponding results are summarized in Table~\ref{tab:HandCraftedMethods}.

\subsection{Dominant Emotion Recognition}
\subsubsection{Traditional Methods}
In the early years, researchers mainly used
%BS: SVM has already been spelt out before - please check coherent introduction and usage of acronyms- some acronyms were never spelt out, some are repeatedly introduced :)
SVM to classify images based on various hand-crafted emotional features.
%SZ: Thanks, I have rewritten it.
%In 2004, \citeauthor{wei2004image}~\cite{wei2004image} propose a new feature representation named weighted line direction-length Vector (WLDLV) by exploring the relationship between line directions and human emotional perception.
%
%SVM is then used as classifier to group images into \textit{static} or \textit{dynamic} based on WLDLV.
%
%\citeauthor{mikels2005emotional}~\cite{mikels2005emotional} defined four positive categories and four negative categories for emotion, which is regarded as theoretical basis of psychology in the most of the classification tasks.
%
\citeauthor{machajdik2010affective}~\cite{machajdik2010affective} combined the features on different levels to generate the final emotion representation.
The experiments are conducted using SVMs on three small datasets using a 5-fold cross validation, and each class is separated against the others in rotation.
The results are reported by true positive rate per class.
%
%
%Generic as well as specific emotion features are utilized to recognize the affective semantics in images.
%
The 1,200 ANPs concept detectors are trained by SVMs, resulting in SentiBank~\cite{borth2013large}, which are the crucial high-level cues for sentiment prediction due to strong co-occurrence relation with sentiments.
As the extensions of SentiBank~\cite{borth2013large}, DeepSentiBank~\cite{chen2014deepsentibank} and MVSO~\cite{jou2015visual} train the detectors for 2,089 and 4,342 English ANPs, respectively, using existing deep architecture like CaffeNet, and then, the sentiment polarity can be inferred.
Using text parsing technology and
%BS:
lexicon-based
sentiment analysis tools, the adjectives can be mapped into ``positive'' or ``negative''; likewise, the polarity of an image is derived.
Even after the emergence of CNNs, SVMs also serve as an essential classifier.
For instance, \citeauthor{ahsan2017towards}~\cite{ahsan2017towards} detected event concepts through a trained CNN model and mapped the visual attributes into specific sentiments based on an SVM classifier.
Besides, hand-crafted art features and CNN features have been combined to generate final representations~\cite{liu2019affective}, which are then input into SVMs for classification.

Inferring the evoked emotion from art paintings has been an interesting research problem in recent years.
Due to the abstract style, recognizing the emotions of art paintings becomes a challenging task.
%
%Based on the low-level features, \citeauthor{zhang2011analyzing}~\cite{zhang2011analyzing} classify the emotions evoked by viewing abstract art paintings into two polarities.
%
Later, considering that an image may be represented in various feature spaces, multiple kernel learning~\cite{zhang2013affective} is employed to capture the different emotional patterns of abstract art.
In the process, the weights of different features can be adjusted automatically, so that the learned feature combination is the most suitable one.
Intuitively, the emotion of paintings is relevant to various characteristics like the painting technique.
Therefore, non-linear matrix completion (NLMC)~\cite{alameda2016recognizing} is introduced as a transductive classifier to model the relations between different latent variables.
This work well imitates the process of inferring emotion from art paintings.
To tackle the scarcity of well-labeled paintings, \citeauthor{lu2016identifying}~\cite{lu2016identifying} proposed an adaptive learning strategy to use the labeled photographs and unlabeled paintings to identify the emotions of paintings.
The differences between the two types of images are considered in the learning process.

% The classification results of different hand-crafted features on small-scale datasets are reported in the first part of Table~\ref{tab:Classification}.
% %
% It can be seen that the most widely used small datasets are IAPSa, Abstract, and ArtPhoto.
% %
% Principles-of-art-based features~\cite{zhao2014exploring} greatly outperforms the combined low-level features proposed in~\cite{machajdik2010affective}, \ie (0.471; 0.357; 0.495) \textit{vs} (0.635; 0.605; 0.609).
% %
% Compared with features designed in~\cite{machajdik2010affective}, principles-of-art-based features~\cite{zhao2014exploring} are more robust to the different arrangements of elements by employing representation derived from different principles.
% %
% In~\cite{rao2016multi}, the individual regions are considered for emotion classification.
% %
% Based on SIFT features, the Bag-of-Visual-Words (BoVW) approach is used to represent the image blocks.
% %
% Benefiting from the multi-scale local features, the performance of classification on three small datasets is further improved (0.699; 0.636; 0.707).

For the same classifier, we compute the average performance of different features as shown in Table~\ref{avg}.
Generally, SVM obtains the best performance in the three classifiers, while $k$NN performs worse than the others except the results on FI8.

\begin{table}[!t]
	\centering\scriptsize
	\setlength\tabcolsep{1.5pt}
	\caption{Average results of different features (PAEF~\cite{zhao2014affective}, Sun~\cite{patterson2012sun}, SentiBank~\cite{borth2013large}, MVSO~\cite{jou2015visual}, pre-trained VGGNet-16~\cite{simonyan2015very}) for the same classifier.}
	\vspace{-5mm}
	\begin{center}
		\begin{tabular}{p{1.5cm}<{\centering} | p{1.5cm}<{\centering} p{1.5cm}<{\centering} p{1.5cm}<{\centering} }
			\toprule
			\textbf{Dataset}	&  \textbf{$k$NN\_Avg}   &  \textbf{NB\_Avg}   & \textbf{SVM\_Avg}     \\  \hline
			Emotion6   &        0.331       &  0.363         &   0.405          \\
			FI2        &        0.721        & 0.738          &  0.793          \\
			FI8        &        0.423        & 0.349          &  0.490          \\
			Flickr        &     0.649        & 0.657          &  0.720          \\
			Instagram     &    	0.614        & 0.636          &  0.691          \\
			Twitter I     &     0.611        & 0.629          &  0.687          \\
			Twitter II    &     0.649        & 0.693          &  0.780          \\
			\bottomrule
		\end{tabular}
		\vspace{-10pt}
	\end{center}
	\label{avg}
\end{table}

%~\cite{kang2018method} construct a dataset that describes the relation between color combination and emotional word, and then recognizes the emotions of paintings by their color spectrum.
\begin{table*}[!t]
	\centering\scriptsize
	\setlength\tabcolsep{1.5pt}
	\caption{Experimental results of learning-based methods on widely-used datasets. The backbone of these methods is replaced with different architectures, including AlexNet (Alex), VGGNet-16 (VGG), ResNet-50 (Res), and Inception-v3 (Inc). The average results of different backbones are reported.}
	\vspace{-5mm}
	\begin{center}
		\begin{tabular}{ p{1.5cm}<{\centering} | p{0.7cm}<{\centering} p{0.7cm}<{\centering} p{0.7cm}<{\centering}  p{0.7cm}<{\centering}p{0.7cm}<{\centering} | p{0.7cm}<{\centering} p{0.7cm}<{\centering} p{0.7cm}<{\centering} p{0.7cm}<{\centering}p{0.7cm}<{\centering} | p{0.7cm}<{\centering}  p{0.7cm}<{\centering} p{0.7cm}<{\centering} p{0.7cm}<{\centering}p{0.7cm}<{\centering}| p{0.7cm}<{\centering}  p{0.7cm}<{\centering} p{0.7cm}<{\centering} p{0.7cm}<{\centering}p{0.7cm}<{\centering}}
			\toprule
			\multirow{2}{1cm}{\textbf{Dataset}} &  \multicolumn{5}{c|}{\textbf{DCNN}~\cite{xu5731visual}}      & \multicolumn{5}{c|}{\textbf{RCA}~\cite{yang2018retrieving}}        & \multicolumn{5}{c|}{\textbf{WSCNet}~\cite{yang2018weakly}}   &       \multicolumn{5}{c}{\textbf{PDANet}~\cite{zhao2019pdanet}}  \\  \cline{2-21}
			& Alex  &  VGG& Res & Inc  & Avg &       Alex  &  VGG& Res & Inc & Avg   &    Alex  &  VGG & Res  & Inc & Avg & Alex  &  VGG & Res  & Inc& Avg   \\ \hline
			%Abstract   & 0.244 & 0.278 & 0.256  & 0.261 & 0.257 & 0.323 &   0.158 & 0.175 & 0.279 &   0.222  & 0.283     &   0.317 & 0.265 & 0.278 & 0.326 \\
			%Artphoto     &0.213  & 0.248 & 0.247 & 0.213 & 0.239  & 0.285  & 0.194  & 0.120 & 0.242 & 0.254 & 0.206  & 0.322 & 0.234 & 0.308 &0.312  \\
			%IAPSa     & 0.142&0.196 & 0.191&   0.124  & 0.183 & 0.194  & 0.152 & 0.155 & 0.233 & 0.186  & 0.194  & 0.179 & 0.198 &0.280  & 0.275 \\
			Emotion6     & 0.489 & 0.483 & 0.532 & 0.546 & 0.513 & 0.512 & 0.516 & 0.546  & 0.559 & 0.533 & 0.463  & 0.514  & 0.551  & 0.495 & 0.506  & 0.488 & 0.517  & 0.569   & 0.487 &  0.515 \\
			FI2         & 0.828  & 0.868 & 0.882  & 0.885 & 0.866  & 0.830  & 0.870 & 0.872 & 0.887& 0.865   & 0.828  & 0.868  & 0.888  & 0.875 &  0.865   & 0.828  & 0.867  & 0.888  & 0.881 &  0.866  \\
			FI8         & 0.576  & 0.614  & 0.660  & 0.668 & 0.630  & 0.559 & 0.641 & 0.668  & 0.678& 0.637  & 0.558  & 0.615  & 0.679  & 0.668 & 0.630   & 0.589  & 0.664  & 0.661  & 0.663 &  0.644  \\
			Flickr       & 0.737 & 0.770 & 0.760 & 0.784  & 0.763  & 0.786  & 0.813 & 0.800 & 0.801 & 0.800 & 0.776 & 0.783  & 0.811  & 0.786 & 0.789  & 0.779  & 0.794  & 0.780  & 0.793 & 0.787 \\
			Instagram       & 0.715 & 0.770 & 0.780 & 0.794  & 0.765  & 0.759 & 0.803 & 0.784 & 0.788  & 0.784  & 0.752 & 0.785  & 0.797  & 0.777 &  0.778  & 0.746 & 0.798  & 0.784  & 0.785 & 0.778  \\
			Twitter I       & 0.791  & 0.811 & 0.829  & 0.810 & 0.810   & 0.802  & 0.834 & 0.825 & 0.821 & 0.821  & 0.772  & 0.805  & 0.826 & 0.814  & 0.804  & 0.782 & 0.698   & 0.833 & 0.846  & 0.790 \\
			Twitter II       & 0.714  & 0.724  & 0.745  & 0.784 & 0.742  & 0.656 & 0.648  & 0.737 & 0.774 & 0.704 & 0.789  & 0.807 &  0.812  & 0.785 & 0.798 & 0.774 & 0.777  & 0.777   & 0.799 & 0.782 \\
			\bottomrule
		\end{tabular}
		\vspace{-10pt}
	\end{center}
	\label{learning_based}
\end{table*}

\subsubsection{Learning-based Methods}
Benefiting from the strong ability of CNNs to extract features, an increasing number of studies~\cite{jindal2015image, campos2015diving, yang2018weakly, zhang2019exploring} design various learning-based methods to recognize image emotions.
In the early studies, a CNN is often directly used as the off-the-shelf tool without any modification.
%
%BS: was FC (fully connected) introduced? Of which net are you talking here?
%SZ: I have revised it.
For example, \citeauthor{xu5731visual}~\cite{xu5731visual} trained two classifiers following the two fully connected (FC) layers (FC7 and FC8) of an existing basic network (AlexNet), respectively.
The experimental results show that the classifier after the FC7 (0.649) layer performs better than that after the FC8 (0.615).
It demonstrates that the 7th layer of CNN characterizes more sentiment information of image than object detection scores in the 8th layer.
To further gain insight about the influence of CNN patterns on visual sentiment analysis, \citeauthor{campos2015diving}~\cite{campos2015diving}~\cite{campos2017pixels} gave a layer-by-layer analysis of a fine-tuned CaffeNet based on both softmax and SVM classifiers.

%Based on the results, it is difficult to claim that any of the two classifiers absolutely performs better than the other.
%
%
%, \ie positive or negative.
%
%To overcome the gap between object classification and emotion recognition, ~\citeauthor{al2019smile}~\cite{al2019smile} pretrain an existing network using Emojis, which can be easily obtained from social media, for image emotion analysis.

\noindent\textbf{Personalized Network.}
With the development of CNNs, researchers have built novel networks for better emotion recognition performance, guided by the theories of art and psychology.
In~\cite{wang2016beyond}, \citeauthor{wang2016beyond} proposed a deep coupled adjective
and noun neural network to recognize positive and negative sentiment from images.
The architecture consisting of two parallel sub-networks (A-net and N-net) can jointly predict the adjectives and nouns of ANPs.
%, which are the important mid-level representations.
%
When ANP labels are unavailable, a mutual supervision is proposed to predict the expected output of each sub-network using a transition matrix that captures the relation between noun and adjective.

\noindent\textbf{Multi-level Features.}
To fully leverage the multi-scale features of image as in~\cite{rao2016learning}, \citeauthor{zhu2017dependency}~\cite{zhu2017dependency} integrated the CNN and RNN architectures.
Specifically, a CNN is used to extract features from different levels, and then, a bidirectional gated recurrent unit (Bi-GRU) captures the dependency among them.
Finally, the two outputs from Bi-GRU are concatenated for emotion classification.
In the learning process, softmax loss and contrastive loss are both used for training the model.
With the
%BS:
contrastive
loss, the features extracted from the images of the same category are enforced to be close to each other, while the features extracted from the images of different categories are enforced to be far away with each other.
This study is the first to model the relations between features on different levels dynamically.% and achieve the best performance at that time, 0.730 on FI dataset.
%\citeauthor{liu2019affective}~\cite{liu2019affective} jointly used several types of interpretable art features and high-level semantic features to recognize the image emotions.
%

\noindent\textbf{Emotional Polarities.}
In Mikels' eight basic emotions, there exist two polarities: \textit{positive} and \textit{negative}.
The emotions in the same polarity are closer to each other---hence, they are highly related.
This characteristic of emotion has been focused upon in several studies.
Based on the triplet loss~\cite{schroff2015facenet}, \citeauthor{yang2018retrieving}~\cite{yang2018retrieving} took the characteristic of polarity into account and designed a sentiment metric loss, in which the quadruplet \{\textit{anchor, positive, related, negative}\} is constructed for learning, where \textit{related} denotes the sample that belongs to the same polarity with the \textit{anchor} but different categories.
By jointly optimizing softmax loss and sentiment metric loss, the architecture can be used for both classification and retrieval tasks.
\citeauthor{he2018emotion}~\cite{he2018emotion} designed a unified architecture consisting of two parts: a sub-network for sentiment polarity classification and a sub-network for specific emotion classification.
With the assisted learning strategy, the results of the polarity can be used as important prior knowledge for more fine-grained emotion analysis.
\citeauthor{yao2019attention}~\cite{yao2019attention} designed emotion-pair loss by considering hierarchical structure in emotions.
Based on the metric learning strategy, the features of the samples from the same polarity will be closer to each other in embedding space.
It is beneficial to rank the images according to the emotional similarity with the given image.

\noindent\textbf{Local Information.}
In complex images, some informative regions may become crucial elements to determine the evoked dominant emotion~\cite{fan2018emotional, cordel2019emotion}.
Therefore, the studies by detecting regions for better recognition performance emerge rapidly.
%focal
\citeauthor{sun2016discovering}~\cite{sun2016discovering}~\cite{yang2018visual} exploited an off-the-shelf
%BS: was "objectness" - please check:
%objectiveness
%SZ: the original paper uses objectness
objectness
tool to generate proposals, and then computed the object and sentiment scores to select top regions from the masses of candidates.
%top $K$ regions
%
The selected regions are aggregated with the whole image for more discriminative representation used in emotion classification.
Based on~\cite{yang2018visual}, \citeauthor{wu2019visual}~\cite{wu2019visual} employed salient object detection model to capture informative regions, and then fed both sub-images and entire images into the network for extracting local and global features.
With the same backbone (VGG-16), \cite{wu2019visual} outperforms \cite{yang2018visual} on all the
%BS:
commonly
used datasets.
It is reasonable to infer that~\cite{wu2019visual} captures more discriminative information by inputting the cropped original images into network.
Obviously, the above methods are time- and computing-consuming when selecting candidates.
Later, WSCNet~\cite{yang2018weakly} is developed to automatically generate an attention map in a single shot based on the response on feature maps, saving considerable amounts of time and computational resources.
Note that each attention map is obtained by computing the weighted sum of the activation for each class.
%
% As shown in Table~\ref{tab:Classification}, the results on all the dataset in~\cite{yang2018weakly} performs better than those in~\cite{yang2018visual} and \cite{wu2019visual}.
% %
% However, in~\cite{rao2019multi}, the performance on FI dataset (\ie 0.755) is better than that (\ie 0.701) in~\cite{yang2018weakly} on the condition of using the same backbone (ResNet-101).
% %
% It is mainly because that~\cite{rao2019multi} exploits the existing object detectors to obtain the region information from multiple layers and adds extra supervision information, \ie label distribution.
In~\cite{fan2017role}, a novel fourth channel, named focal channel, is added in neural networks by taking the focal object mask of the image or the saliency map as input.
By encoding the local information for sentiment representation, it is shown that negative sentiment is mainly evoked by the focal region and hardly influenced by context, whereas positive sentiment is decided by both focal region and context.
%
%\citeauthor{rao2019multi}~\cite{rao2019multi} combined the region information from multiple levels, the emotional classification performance is improved significantly.
%
A Sentiment Network with visual Attention (SentiNet-A) is proposed in~\cite{song2018boosting}, where the attention distributions of spatial regions are generated.
The saliency map is then derived from a multi-scale fully convolutional network (FCN) to refine the attention distribution.
Recently, how to capture the emotional relation between different regions is a hot topic.
\citeauthor{zhang2019object}~\cite{zhang2019object} modeled the correlation between object semantics in different image regions to infer the image sentiment based on Bayesian networks.
%i
A multi-attentive pyramidal model is proposed in~\cite{he2019multi} to extract local features at various scales, and then, a self-attention mechanism is employed to mine the relations between features of different regions.

\noindent\textbf{Knowledge of Other Fields.}
Studying emotion recognition can also utilize knowledge from other fields.
Considering the correlation between aesthetics and emotion of images, \citeauthor{yu2019towards}~\cite{yu2019towards} designed a novel unified  aesthetics-emotion hybrid network (AEN) to simultaneously conduct image aesthetic assessment and emotion recognition.
%
%Besides, both hand-crafted art features and CNN features are also
%
Inspired by the emotion of the generation process in brain, \citeauthor{zhang2019another}~\cite{zhang2019another} developed a multi-subnet neural network to simulate the generation of specific emotional signals and the process of signal suppression in brain neurons.

% We list the representative studies on dominant emotion recognition and the corresponding experimental results on different datasets in Table~\ref{tab:Classification}.
% %
% One can observe that the learning methods based on CNNs generally perform better than those based on hand-crafted features.
% %
% The architectures with deeper CNN models are generally able to learn more discriminative features.
% %
% Within hand-crafted features, high-level features show better performance than low-level and mid-level ones when used for different tasks, including classification and retrieval.
% %
% For abstract paintings, low-level features (such as Elements and IttenColor) that focus on color, texture, \etc perform well, while mid-level features (such as Principles) are effective for artistic photos.

\noindent\textbf{Quantitative Comparison of Representative Deep Methods.}
As shown in Table~\ref{learning_based}, we conduct experiments to fairly compare four representative learning-based methods, including DCNN~\cite{xu5731visual}, RCA~\cite{yang2018retrieving}, WSCNet~\cite{yang2018weakly}, and PDANet~\cite{zhao2019pdanet}. We replace the original backbone with four different backbones to evaluate the effectiveness and robustness, including AlexNet~\cite{krizhevsky2012imagenet}, VGG-16~\cite{simonyan2015very}, ResNet-50~\cite{he2016deep}, and Inception-v3~\cite{szegedy2016rethinking}.
It is observed that the robustness of different methods is different.
Compared to WSCNet and PDANet, RCA is more robust when using different architectures as backbones, because the results of RCA on each dataset fluctuate less than other methods, especially on Flickr and Emotion6.
In RCA, the final image representation contains features from multiple layers, leading to richer information, of which distinctive ability does not decrease dramatically when using shallower networks.
By contrast, the methods that are only based on features of the final layer are more sensitive to the depth of the used backbones.
Generally, the results of recent studies, such as RCA, WSCNet, and PDANet, are better than that of DCNN, which does not contain specialized components designed for emotion classification.
Under the common experimental settings, including the same input size, initialization, backbone, \etc, the overall results of RCA, WSCNet, and PDANet do not have distinct disparity.
For different datasets, the methods that obtain the best performance are different.
For example, for one small-scale dataset Twitter II (only 603 images), the methods (WSCNet and PDANet) considering local informative features perform better than others.

\subsection{Personalized Emotion Prediction}

\citeauthor{yang2013user}~\cite{yang2013user} first proposed to predict emotion of social images for individuals based on user interest and social influence. The user interest is modeled by considering both text and images, the emotions of which are predicted by constructing a personalized dictionary and clustering basic color features. The social influence is measured by the emotion similarity of different users towards the same microblog. The weights of user interest and social influence are obtained by mining users’ historical behaviors. Later, \citeauthor{rui2017joint} extended the weighting strategy with a probabilistic graphical model~\cite{rui2017joint}. Latent from the user's historical behaviors, a set of parameters in the graph model are used to estimate the importance of content and influence. However, there are some limitations of these methods. First, the extracted visual features are very simple, which cannot well reflect the visual content. Second, several important factors are not considered, such as the temporal evolution. Third, the higher-order correlations among users and images are not well modeled.

In~\cite{zhao2016predicting,zhao2018predicting}, \citeauthor{zhao2016predicting} made several improvements to address these issues when predicting the personalized emotions (see Figure~\ref{fig:SubjectivityExamples} (b)) of a specified user after viewing an image, associated with online social networks. Different types of factors that may influence the emotion perception are considered: the images' visual content, the social context related to the corresponding users, the emotions' temporal evolution, and the images' location information. Rolling multi-task hypergraph learning is presented to jointly combine these factors. Each hypergraph vertex is a compound triple $(u,x,S)$, where $u$ represents the user, $x$ and $S$ are the current image and the recent past images, termed as `target image' and `history image set', respectively. Based on the 3 vertex components, different types of hyperedges are constructed, including target image centric, history image set centric, and user centric hyperedges. Visual features (Gist, Elements, Attributes, Principles, ANP, and Expressions) in both the target image and the history image set are extracted to represent visual content. User relationship is exploited from the user component to take social context into account. Past emotion is inferred from the history image set to reveal temporal evolution. Location is embedded in both the target image and the history image set. Semi-supervised learning is then conducted on the multi-task hypergraphs to classify personalized emotions for multiple users simultaneously.
%The average F1 of emotion classification on the IESN dataset is 0.582.
%The average performance of emotion classification on the IESN dataset measured by precision, recall and F1 are 0.49, 0.72 and 0.58, respectively.
%The experiments are conducted on the IESN dataset.

% \begin{figure}[!t]
% 	\begin{center}
% 		\includegraphics[width=0.95\linewidth]{image_ldl2.pdf}
% 		\caption{Images and the corresponding label distributions. They are selected from (a) Flickr\_LDL and (b) Abstract. The different colors represent different emotion categories.}
% 		\label{fig:Image_LDL}
% 	\end{center}
% \end{figure}

\subsection{Emotion Distribution Learning}
Label distribution learning (LDL)~\cite{geng2013facial} is used to model the relative importance of each category for an image.
The sum of the probability on each label in discrete space is 1.
It is usually used to solve the ambiguity of emotion in discrete label space.%, as shown in Fig.~\ref{fig:Image_LDL}.
%
%We summarize the representative studies and corresponding experimental results in Table~\ref{tab:Distribution}.

\citeauthor{peng2015mixed}~\cite{peng2015mixed} constructed the Emotion6 database that is annotated with probability distribution.
SVR, CNN, and CNN regression (CNNR) are employed as the emotion regression model.
Particularly, different SVR and CNNR models are trained for each category, while the CNN is trained for all categories by changing the number of output neurons to the number of emotion categories.
\citeauthor{zhao2015predicting}~\cite{zhao2015predicting, zhao2018discrete} modeled the emotion distribution prediction as a shared sparse learning (SSL) problem.
The input is the combination of different types of features, and the objective is iteratively optimized by reweighted least squares.
Later in~\cite{zhao2017approximating}, the weighted multi-modal shared sparse learning (WMMSSL) is proposed, in which the weight of different features can be learned automatically.
Based on the conditional probability neutral network (CPNN), in~\cite{geng2013facial}, BCPNN~\cite{yang2017learning} is proposed by representing the image label with binary encoding rather than the general signless integers.
Besides, ACPNN is developed based on BCPNN by adding noises to the ground truth label.
With this strategy, the emotion distribution is augmented, which benefits training more robust models.
\citeauthor{zhao2017learning}~\cite{zhao2017learning} proposed a Weighted Multi-Modal Conditional Probability Neural Network (WMMCPNN) to explore the optimal combination coefficients of different types of features.
In~\cite{yang2017joint}, \citeauthor{yang2017joint} designed a unified framework to optimize the Kullback-Leibler (KL) loss and softmax loss, simultaneously.
Besides, considering the lack of manually annotated emotion distribution in some datasets, a scheme that converts the single emotion into a probability distribution is proposed in this study.

Considering the co-occurrence and mutual exclusion of some emotions, it is important to model the relation of different emotional labels when predicting the probability labels.
\citeauthor{he2019image}~\cite{he2019image} used Graph Convolutional Networks (GCN) to model the label relationship for label distribution prediction.
In detail, the GloVe-300 word embeddings of emotions are input into GCN as nodes, and the relation of different labels is computed using the probability of co-occurrence of two emotions.
\citeauthor{liu2018structured}~\cite{liu2018structured} integrated low-rank and inverse-covariance regularization terms into one framework for emotion distribution learning.
The low-rank regularization term is used to learn low-rank structured embedding features, while the inverse-covariance regularization term can ensure the structured sparsity of regression coefficients.
To fully employ the polarity and character of the intensity in emotions, structured and sparse annotations are leveraged to learn an emotion label distribution in~\cite{xiong2019structured}.

%In the continuous emotion space, the emotion distribution is represented in VA space.
%
In the continuous emotion space, \citeauthor{zhao2017continuous}~\cite{zhao2017continuous} modeled the continuous distribution with a Gaussian mixture model, in which the parameters can be estimated by the expectation-maximization algorithm. Shared Sparse Regression (SSR) is introduced as the learning model by assuming that the test feature and test parameters can be linearly represented by the training features and training parameters but with shared coefficients. To explore the task relatedness, multi-task SSR is further proposed to simultaneously predict the parameters of different test images by using proper shared information across tasks.
% Gist, Elements, Attributes, Principles, ANP, and 4,096-dimensional deep features from AlexNet are extracted as visual features. Experiments are conducted on a subset of IESN, which consists of 18,700 images each with more than 20 VA labels. The average KL divergence of multi-task SSR using ANP is 0.436.

\subsection{Learning from Noisy Data or Few Labels}

\textbf{Few-shot or Zero-shot Learning.} As stated in Section~\ref{ssec:Challenges}, few/zero shot learning and unsupervised/weakly supervised learning are two possible solutions to address the label absence challenge. Few/zero shot learning refers to a specific machine learning, where a model is learned based on very few or even no labeled examples~\cite{zhan2019zero}. Although humans can learn through only a small number of samples, it is difficult for machine to do so. Conventional methods usually construct a shared space for both seen and unseen classes. For seen classes, the space is learned based on the correspondence between the seen images and their labels. Relying on the side information (\eg attributes), the unseen classes are first related to the seen classes and then mapped to the common space based on the cross-modality similarity between visual features and class semantic representations. The existence of the affective gap makes it difficult to compute this similarity.

\citeauthor{wang2019robust}~\cite{wang2019robust} proposed an emotion navigation framework using auxiliary noisy data and employed the few-shot precise samples as the prototype center to guide noisy data clustering. \citeauthor{zhan2019zero}~\cite{zhan2019zero} proposed an affective structural embedding framework, which constructs an intermediate embedding space using  ANP features for zero-shot emotion recognition. In addition, an affective adversarial constraint is introduced to select the embedding space that simultaneously preserves the affective structural information and retains the discriminative capacity.

\textbf{Unsupervised/Weakly-supervised Learning.} Unsupervised learning aims to find previously unknown patterns in a dataset without pre-existing labels. Two main methods are cluster analysis and principal component analysis. How to automatically determine the number of clusters is a key challenge in clustering. Differently, \citeauthor{wang2015unsupervised}~\cite{wang2015unsupervised} exploited the relations among visual content and relevant textual information for unsupervised sentiment analysis of social images. This method relies on the accompanying text of social images. On the one hand, the text may be incomplete and noisy. On the other hand, there may be no available text. In such cases, how to conduct unsupervised analysis is worth studying.

For social images, a more practical scenario is that they are weakly and noisily labeled~\cite{wang2018visual,wu2017reducing,chen2018predicting,wei2020learning}.
%Based on a large amount of free data, \citeauthor{panda2018contemplating}~\cite{panda2018contemplating} constructed a new database, named WEBEmo, which covers a large variety of concepts.
%
%The model trained by the new dataset can be well generalized to other dataset with little drop on classification accuracy.
%
Considering that the images of the VSO dataset are weakly labeled with noises, \citeauthor{wang2018visual}~\cite{wang2018visual} estimated the noise matrix to reweight the softmax loss that can compensate the degeneration of classification performance resulting from the noisy labels. Retrained by reweighting the loss, the learned model is more discriminative for emotional images. \citeauthor{wu2017reducing} proposed to refine the weakly labeled dataset based on the sentiments of ANPs and provided tags~\cite{wu2017reducing}. The images are removed if the sentiments of ANPs and tags are contradicting each other and if the numbers of positive tags and negative tags are equal. The remaining images are relabeled with the dominant sentiment of the tags. Using the refined dataset, a better performance can be obtained. \citeauthor{chen2018predicting} employed a probabilistic graphical model to filter out the label noise~\cite{chen2018predicting}. \citeauthor{wei2020learning}~\cite{wei2020learning} proposed to train a joint text and visual embedding to reduce noise in the webly annotated tags by text-based distillation. Designing an effective strategy that can better refine the dataset or filter the label noise is expected to improve the performance.

\textbf{Domain Adaptation/Generalization.} Domain adaptation studies how to transfer the models trained on a labeled source domain to another sparsely labeled or unlabeled target domain. One direct solution is to translate the source images to an intermediate domain that is indistinguishable from the target images using GANs~\cite{goodfellow2014generative,zhu2017unpaired,zhao2021madan}. Meanwhile, the source labels should be preserved. Some existing unsupervised domain adaptation methods on AICA are based on this intuition. \citeauthor{zhao2018emotiongan}~\cite{zhao2018emotiongan} studied the domain adaptation problem in emotion distribution learning. They develop an adversarial model, termed EmotionGAN, by alternately optimizing the GAN loss, semantic consistency loss, and regression loss. The semantic consistency loss guarantees that the translated intermediate images preserve the source labels. Since traditional GANs are unstable and prone to failure~\cite{zhu2017unpaired}, the cycle-consistent GAN (CycleGAN) was designed. Based on CycleGAN, \citeauthor{zhao2019cycleemotiongan} enforced semantic consistency when adapting the dominant emotions without requiring aligned image pairs~\cite{zhao2019cycleemotiongan,zhao2021emotional}. \citeauthor{he2019deep} proposed a discrepancy-based domain adaptation method~\cite{he2019deep}. Both marginal and joint domain distribution discrepancies at fully-connected layers are reduced by minimizing the joint maximum mean discrepancy. Without generating an intermediate domain, this method aims to extract more transferable features.

% Both, marginal and joint domain distribution discrepancies at fully-connected layers, are reduced by minimizing the joint maximum mean discrepancy. Without generating an intermediate domain, this method aims to extract more transferable features.

All the above methods focus on a single-source scenario. However, in practice, the labeled data may be collected from multiple sources with different distributions. Simply combining the multiple sources into one source and performing single-source domain adaptation may lead to suboptimal solutions. In~\cite{lin2020multi}, \citeauthor{lin2020multi} studied multi-source domain adaptation for binary sentiment classification of images. Specifically, a multi-source sentiment generative adversarial network (MSGAN) is designed to find a unified sentiment latent space where the source images and target images share a similar distribution. MSGAN includes three pipelines: image reconstruction, image translation, and cycle reconstruction. The results demonstrate that exploring the complementarity of multiple sources can improve the adaptation performance to a large margin as compared to best single-source adaptation methods.

Differently, \citeauthor{panda2018contemplating}~\cite{panda2018contemplating} studied the domain generalization problem of AICA to overcome dataset bias. A weakly-labeled large-scale emotion dataset is constructed by collecting images from a stock website to cover a wide variety of emotion concepts. A simple yet effective curriculum guided training strategy is proposed to learn discriminative emotion features, which demonstrate better generalization ability than the existing datasets.

\section{AICA Based Applications}
\label{sec:Applications}

With the booming development of AICA, the related application has been or will be on the agenda in different directions, including opinion mining, business intelligence, psychological health, and entertainment assistant, to name but a few in more detail.% in the following.

\subsection{Opinion Mining}
Nowadays, an increasing number of people use images to express their viewpoints or attitudes towards some events.
Based on the analysis of these shared images, we can infer the emotions of the different users, including uploaders and commentators.
Furthermore, we can conjecture their attitudes towards the specific events or products. In~\cite{zhao2016predicting}, the different types of factors, including visual content, social context, temporal evolution and location influence, are modeled using a hypergraph model to iteratively optimize the personal social image emotion prediction.
Furthermore, various virtual groups are formed according to the interests or backgrounds of users.
Analyzing group-based emotions will contribute to predicting the tendency of the society.
Based on the above technologies, we can imagine that the understanding of social image emotion can be used in public opinion analysis and related applications.

In special domains like product comments, the experience of users has been investigated and evaluated based on emotions from uploaded images.
In~\cite{truong2017visual}, \citeauthor{truong2017visual} conducted visual sentiment analysis for better understanding of review images about different products, services, and venues.
In the process, both user and items factors are taken into account.
\citeauthor{ye2019visual}~\cite{ye2019visual} jointly employed a visual and textual classification to analyze the sentiment of the product reviews.
Besides, a dataset named Product Reviewes-150K (PR-150K) is constructed.
In~\cite{hassan2019sentiment}, \citeauthor{hassan2019sentiment} analyzed the sentiments evoked from disaster-related images by taking into account people's opinions, attitudes, feelings, and emotions.
The study sets a baseline for the future research in disaster-related images sentiment analysis.
Therefore, it is significant to mine the positive or negative aspects for opinion of the users by analyzing the emotions of related images.

\subsection{Psychological Health}
%Based on the theory of psychology,  . It postulates that these illnesses have an onset in which a cognitive evaluation initiates a sequence of unconscious transitions yielding a basic emotion.
%
%This emotion is appropriate for the situation but inappropriate in its intensity.

With the popularity of the social media, people share their mood on the Internet rather than with their real friends.
For a user that shares negative information continuously, it is necessary to further track her/his mental status to prevent the occurance of psychological illness and even suicide.
\citeauthor{guntuku2019twitter}~\cite{guntuku2019twitter} revealed how a twitter profile and post images reflect depression and anxiety.
In~\cite{lin2014psychological}, an automatic stress detection model is proposed for social web users by analyzing the emotional content of multi-modal microblog data.
Based on the model, we can further design subsequent decompressing services for users, including playing some smoothing music, playing some funny videos, and providing some forms of exercises, \etc.

In the field of psychology, affective images are employed to conduct some studies.
For instance, IAPS~\cite{lang1997international} is a database of images constructed to provide a standardized set to evoke a target emotion in people for studying psychological status.
Each image has listed the average ratings of the elicited emotions, and these ratings can be used for various research directions in psychological theories.
In~\cite{bao2014thupis}, a new image system named Tsinghua psychological image system (ThuPIS) is built based on the Minnesota multiphasic personality inventory (MMPI)~\cite{helmes1993perspective}, which is a famous personality diagnosis tool for clinical mental health.
The system can be applied to support the new psychological test for monitoring the mental health of humans.

\subsection{Business Intelligence}
Images play an essential role in conveying the business information, so selecting the images with proper emotions can benefit the development of a business.
For example, most advertisements are presented using visual content to evoke strong emotional stimulus in viewers.
%
%As shown in Fig.~\ref{fig:advertise}, these advertisement photos can evoke strong emotional stimulus in viewers.
%BS are you sure you are allowed to use these images? Copyright might be an issue?
%SZ: these images are obtained from google search. We can give the original references. Is that enough?
% There are more such images in the paper "Interpreting the Rhetoric of Visual Advertisements" (https://ieeexplore.ieee.org/stamp/stamp.jsp?tp=&arnumber=8869939). So this might be not a big problem.
Consumer research~\cite{garg2007influence} has proven that emotions can affect the process of decision making.
A well-designed advertisement can attract people's attention and evoke positive emotions in viewers, so that a desire of purchasing will be produced when viewing an accordingly tailored advertisement.
\citeauthor{holbrook1984role}~\cite{holbrook1984role} investigated the role of emotion in advertising.
Specifically, they distinguish emotion from other types of consumer responses, and study the emotion generating process from the emotional content in the advertisements.
Besides, suggestions are put forward for the design of advertisement considering emotional elements in the future.
\citeauthor{poels2006capture}~\cite{poels2006capture} reviewed and updated the measuring  methods for emotions in advertising, and further discuss their applicability.
Finally, the influence of emotions on the effectiveness of advertising is investigated.

% \begin{figure}[!t]
% 	\begin{center}
% 		\includegraphics[width=0.95\linewidth]{advertisement.pdf}
% 		\caption{Examples of advertisement photos. Each image can create a pleasant atmosphere using the context. For example, viewers will feel warmth and contentment from (a) and (c), while (b) and (d) may elicit awe emotions.}
% 		\label{fig:advertise}
% 	\end{center}
% \end{figure}

In the filed of tourism, emotion is an important element that cannot be ignored for evaluating the overall experience of a trip~\cite{hosany2010measuring}.
By analyzing the uploaded travel photos in the social networks, the relations among motivation, image dimension, and emotional qualities of places are explored in~\cite{pan2014travel}.
The paper reveals that the natural resources, including ``flora and fauna'', ``countryside'', ``beaches'', \etc, are always associated with the feelings of ``arousing'' and ``pleasant'' for the specific destination.
Besides, the travel photos taken in a long shot, at eye-level, with stark density level can elicit happiness feelings.
These findings can guide to exhibit the more attractive travel photos on some specific platforms, so as to initiate successful marketing efforts and promote the booming of tourism.
In~\cite{toyama2016categorization}, a survey is conducted on the emotional experience of tourists by distributing self-administered questionnaires.
The emotion feedback (\textit{arousal} and \textit{pleasure}) for each destination is plotted on a corresponding two-dimensional grid.
\citeauthor{hosany2013patterns}~\cite{hosany2013patterns} empirically investigated the patterns of emotional response from tourists and discussed the relationship between these emotional patterns and the consumption satisfaction.
Five different emotional response patterns (\textit{delighted}, \textit{unemotional}, \textit{negative},
\textit{mixed}, and \textit{passionate}) are derived by cluster analysis based on a four-dimensional emotional space defined by love, surprise, joy, and unpleasantness.
It is reported that these five patterns are different in the satisfaction level and the intention of recommendation. In the future, based on a more fine-grained emotion analysis, we can construct personalized destination recommendation systems for users who intend to have different travel experiences automatically.

\subsection{Entertainment Assistant}
Nowadays, the standard of entertainment has treated the emotion as a crucial element that can decide the entertainment experience~\cite{tan2008entertainment}.
Simultaneously, the emotion can also be used to evaluate the experience of entertainment.
For instance, emotions can be regarded as the medium that links different modalities of data, such as image and music.
In~\cite{xing2015emotion}, an emotion-driven cross-media retrieval system is designed based on differential and evolutionary-support vector machine (DE-SVM).
The system can achieve the retrieval between Chinese folk music and Chinese folk image based on their involved emotions.
\citeauthor{chen2014object}~\cite{chen2014object} and \citeauthor{zhao2020emotion}~\cite{zhao2020emotion} designed a system that computes the emotional similarity between music and images.
With this system, users can generate the mood-aware music slide shows from their personal album photos.

Emotions in comics play a crucial role in attracting people.
The reason why Indonesian readers widely accept Japanese comics has been investigated in~\cite{ahmad2012emotion}.
The report indicates that the comics are not only an entertainment for them, but also a significant life experience, in which emotion is an important element.
Therefore, we should take into account the evoked emotions when measuring comics.
In~\cite{she2019learning}, a large-scale comics dataset is constructed, in which the images are labeled with the emotions defined in Mikel's wheel.
With more attention paid to AICA in the entertainment domain, we can establish a  conversation with chatbots based on various types of images rather than only based on text.

\section{Future Directions}
\label{sec:FutureDirections}

% \begin{figure}[!t]
% \begin{center}
% \includegraphics[width=0.9\linewidth]{ImageContent.pdf}
% 		\caption{Necessity examples of image content. Subtle analysis of image content for AICA: the laugh of a lively child (left) and that of an evil ruler (right) may evoke different emotions-- amusement and anger or ``schadenfreude''.}
% 		\label{fig:ImageContent}
% \end{center}
% \end{figure}

Although remarkable progress has been made on affective image content analysis (AICA), there are still several open issues and directions that are worth investigating by jointly considering the efforts from different disciplines, such as psychology, cognitive science, multimedia, and machine learning.

\subsection{Image Content and Context Understanding}

As emotions may be directly evoked by the image content in viewers, accurately analyzing what is contained in an image can significantly improve the performance of AICA. As stated in Section~\ref{sec:Features}, there are different kinds of emotion features. Although the deep ones generally outperform the hand-crafted ones, it is unclear whether combining hand-crafted ones with deep ones can boost the performance. If yes, how to effectively fuse them? Further, the correlation between deep features and specific emotions is unclear, while the hand-crafted features---especially mid-level and high-level ones---are more understandable. Using hand-crafted features to guide the generation of interpretable deep ones is an interesting topic. Sometimes, we even need subtle analysis of image contents. For example, we may feel ``happy'' about beautiful flowers; but, if the flowers are placed in a funeral, we possibly feel ``sad''. If an image is about the laugh of a lovely child, it is more likely that we feel ``amused''; but if it is about the laugh of a known evil ruler or criminal, we may feel ``angry''.%, as shown in Fig.~\ref{fig:ImageContent}.
Constructing a large-scale repository and collecting sufficient corresponding images can help to solve this problem. As shown in Fig.~\ref{fig:ImageContext}, the context of an image is also very important. Multi-modal emotion recognition would make more sense, such as textual-visual data~\cite{you2016robust,chen2018predicting} and audio-visual data~\cite{hossain2019emotion}. One key challenge is how to fuse the data of different modalities.

\subsection{Viewer Contextual and Prior Knowledge Modeling}

The contextual information of a viewer watching images can significantly influence the emotion perception. The same viewer can experience different emotions for the same image depending on the context, such as climate, time, and social context~\cite{zhao2016predicting}. Incorporating these important contextual factors can be expected to boost the performance. Using probabilistic graph or hypergraph models to represent the complex correlations of different factors is demonstrated to be feasible~\cite{yang2014your,wang2015modeling,zhao2016predicting}. We may further try to model these factors by more recent graph convolutional networks~\cite{kipf2017semi} and hypergraph neural networks~\cite{feng2019hypergraph}.

The prior knowledge of viewers, such as gender and personality, may also influence the emotion perception. For example, an optimistic viewer and a pessimistic viewer may have totally different emotions about the same image. \citeauthor{wu2015understanding} investigate the influence of user demographics, including gender, marital status, and occupation, as related to the emotion perception of social images~\cite{wu2015understanding,wu2017inferring}. Besides the visual content, temporal correlation, and social correlation, user demographics are also incorporated as factor functions in a factor graph model. The results show that user demographics can indeed improve the overall emotion classification performance. However, the collected prior knowledge on social networks may be inaccurate. How to automatically filter the noisy ones has not been investigated.

\subsection{Learning from Noisy Data or Few Labels}

\textbf{Few-shot or Zero-shot Learning.}
There are some limitations of current few-shot/zero-shot learning methods for AICA~\cite{wang2019robust,zhan2019zero}. First, not all seen images are helpful in generating the embedding space. How to automatically select the representative images to generate better embedding space is unclear. Second, the embedding process may result in information loss, and the cross-modality similarity cannot make full use of the data distribution. We may consider synthesizing reliable samples for the unseen classes based on the estimated distribution. With the success of Generative Adversarial Networks (GANs)~\cite{goodfellow2014generative}, such an idea would bear genuine potential.

\textbf{Domain Adaptation/Generalization.} In practice, we might have a few labeled target images. In such cases, the domain adaptation task becomes a semi-supervised scenario. One interesting problem is how many labeled target images are required at least to achieve or even outperform the results fully trained on the target domain. Besides the semi-supervised domain adaptation, some other challenging problems include heterogeneous domain adaptation where the label space is different between the source and target domains, open set domain adaptation where both source and target domains contain images that do not belong to the classes of interest, and category shift domain adaptation where the categories from different sources might be different.

While the target images (although without labels) are available in domain adaptation, \ie the target images are accessible during the training process, domain generalization learns a model without accessing any target image~\cite{muandet2013domain}. To enrich the generalization ability, one possible solution is to randomize the labeled source images to a sufficient number of domains
in the training stage using domain randomization~\cite{tobin2017domain}, and then, the target domain belongs to the randomized domains to a large extent and thus, the models trained on the randomized domain can well adapt to the target domain.

\subsection{Group Emotion Clustering}

Simply recognizing the dominant emotion for an image is too generic, while predicting personalized emotion for each user is too specific. Since some groups or cliques of users, who share similar tastes or interests and have similar background, are more likely to respond  similarly to the same image, it would make more sense to predict emotions for these groups or cliques. Analyzing the user profiles provided by each individual to classify users into different types of groups based on gender, backgrounds, tastes, interests, and so on may provide a feasible solution.

Current research on group emotion mainly focuses on recognizing the emotions of the groups of people contained in an image attending a wide variety of social events~\cite{dhall2018emotiw,guo2018group}. Affective image analysis for groups of people, \ie recognizing the induced emotions of the groups, has not been explored yet. Group emotion recognition plays an important role in recommendation. For example, for the people in the same group, if one is interested in a specific product, the others are more likely to accept it.

% \begin{figure}[!t]
% \begin{center}
% \includegraphics[width=0.95\linewidth]{Implicit.pdf}
% \caption{Direct AICA (solid) vs.\ implicit AICA (dashed). The emotion learning model is usually different for these two approaches.}
% \label{fig:Implicit}
% \end{center}
% \end{figure}

\subsection{Viewer-Image Interaction}
Besides the direct analysis of image content, we may also record and analyze the viewers' audiovisual or physiological responses when watching the image (such as facial expressions, or electroencephalogram signals), which is often called implicit emotional tagging.%, as shown in Fig.~\ref{fig:Implicit}.
Current methods mainly focus on videos~\cite{soleymani2011multimodal,joho2011looking,zhao2013video,koelstra2013fusion} for its relative emotional consistency temporally. Exploring viewers' responses for implicit emotion analysis of images is still a largely open topic of research. Jointly modeling both image content and viewers' responses may better bridge the affective gap and result in superior performance. In practice, some data may be missing or corrupted. For example, some physiological signals are not successfully captured. In such cases, how to deal with missing data should be considered.

As explained in Section~\ref{sec:Introduction}, the physiological responses are either difficult to capture or easily suppressed. In real-world applications, even if there are no physiological responses, jointly exploring the privileged modality during training might also lead to better performance than using the image modality only.

\subsection{Novel and Real-world AICA-based Applications}

With the availability of large-scale datasets and improvements in machine learning, especially in deep learning, the AICA performance will be significantly boosted. Therefore, we foresee the coming of an emotional intelligence era with more AICA-based real-world applications. For example, in online fashion recommendation, intelligent costumer services, such as customer-image interaction, can provide better experience to customers. In advertisement, generating or curating images that can evoke expected emotions strongly can attract more attention. One preliminary image adjustment system is implemented in~\cite{wang2013affective}. Given an input image and an affective word, the system can adjust image color to meet the desired emotion. Only the color information is changed, which may be insufficient in applications. \citeauthor{peng2015mixed} instead proposed to modify the evoked emotion distribution of the given source image towards that of the target image by changing color and texture related features~\cite{peng2015mixed}. We believe that GAN-based adversarial models are possibly more suitable to generate affective images. In art theory, we can understand how artists express emotions through their artworks. The principles can guide the affective image generation. The generated synthetic images can in turn improve the AICA results through domain adaptation. In education, the images with enriched emotions can help children to better learn and understand. Certainly, many more exciting applications will be coming up soon.

\subsection{Efficient AICA Learning}

There are three factors that attribute to the success of deep learning: increased computing capacity, deep complex models, and sufficient labeled data. However, these factors may be unavailable for edge devices such as mobile phones which are widely used in our daily life but have limited power, memory, and computing capacity. Therefore, designing specialized and efficient ``green'' deep learning models is required. Efficient model design has been actively studied in computer vision. Some efficient representation methods include auto channel pruning, student-teacher network approaches, neural network and hardware accelerator co-design, auto mixed-precision quantization, optimal neural architecture search, \textit{etc.}

To the best of our knowledge, the efficiency problem has not been well studied in AICA. Extending existing methods in computer vision to the AICA task by incorporating its speciality (\eg emotion hierarchy) is a simple but effective solution. It would make more sense if the on-device training models can learn online with incremental data.

\subsection{Benchmark Dataset Construction}

The datasets adopted in existing AICA studies are mainly well-labeled small-scale ones (\eg IAPSa~\cite{mikels2005emotional}) or large-scale ones with labels obtained by a keyword searching strategy (\eg IESN~\cite{zhao2016predicting}). While there are not enough training samples in the former ones, the label quality of the automatic annotations cannot be guaranteed in the latter case. Creating a large-scale and high-quality dataset, like the ImageNet in computer vision, can significantly advance the development of AICA. One possible solution is to exploit online systems and crowd-sourcing platforms to invite/attract large numbers of viewers with a representative spread of backgrounds to annotate their personalized emotion perceptions of images together with the contextual information on their emotional responses. Personalized emotion annotation would better accord with the subjectiveness of emotions. Further, from the personalized emotions, we can obtain both the dominant emotion and emotion distribution. Collecting the social media users' interaction with images, \eg, likes, comments, together with their spontaneous responses, \eg, facial expression, where possible, can provide more information to enrich affective datasets. To facilitate the applicability of AICA in practice with different emotion requirements, employing a hierarchical model (\eg Parrott~\cite{parrott2001emotions}) with emotion intensity is a good choice.

\section{Conclusion}
%attempted
This article attempted to provide a comprehensive survey of recent developments on affective image content analysis (AICA) over the last two decades. Obviously, it cannot cover all the literature on AICA, and we focused on a representative subset of the latest methods. We summarized and compared the widely employed emotion representation models, available datasets, and the representative works on emotion feature extraction, learning methods, and AICA-based applications. Finally, we discussed some open issues and potential research directions in AICA. Although deep learning-based AICA methods have achieved remarkable progress recently, an effective, efficient, and robust AICA algorithm that can achieve satisfying performance under unconstrained conditions is yet to be designed. With the rapid development of deep understanding of emotion evocation in brain science, accurate emotion measurement in psychology, and novel deep learning network architectures in machine learning, we believe that AICA will continue to be an active and promising research topic for a long time.

\noindent{\textbf{Acknowledgements}: This work is supported by the National Natural Science Foundation of China (Nos. 61701273, 61876094, U1933114, 61925107, U1936202), the National Key Research and Development Program of China Grant (No. 2018AAA0100403), the Natural Science Foundation of Tianjin, China (Nos.20JCJQJC00020, 18JCYBJC15400, 18ZXZNGX00110), and Berkeley DeepDrive.}

\ifCLASSOPTIONcaptionsoff
\newpage
\fi

\bibliographystyle{IEEEtranN}\scriptsize%\footnotesize%\small
\bibliography{EmotionSurvey}  % sigproc.bib is the name of the Bibliography in this case

%\clearpage
\vspace{-30pt}
\begin{IEEEbiographynophoto}{Sicheng Zhao}
(SM'19) received the Ph.D. degree from Harbin Institute of Technology, Harbin, China, in 2016. He worked as a Research Fellow at Tsinghua University from 2016 to 2017 and at University of California, Berkeley from 2017 to 2020. His research interests include affective computing, multimedia, and computer vision.
\end{IEEEbiographynophoto}
\vspace{-30pt}

\begin{IEEEbiographynophoto}{Xingxu Yao}
received the Master degree from Nankai university, Tianjin, China, in 2021.  His research interests include computer vision, affective computing, and multimedia.
\end{IEEEbiographynophoto}
\vspace{-30pt}

\begin{IEEEbiographynophoto}{Jufeng Yang}
received the Ph.D.\ degree from Nankai University, Tianjin, China, in 2009. He is currently a full professor in the College of Computer Science, Nankai University and was a visiting scholar with the Vision and Learning Lab, University of California, Merced, USA, from 2015 to 2016. His recent interests include affective computing, image retrieval, fine-grained classification, and medical image recognition.%He is a member of CCF, ACM and IEEE. He won CSC Scholarship, IBM Chinese Excellent Students Scholarship and Zhou En-Lai Scholarship in the last decade. He served as an organizing committee chair of CCCV 2017 and CVM 2017.
\end{IEEEbiographynophoto}
\vspace{-30pt}

\begin{IEEEbiographynophoto}{Guoli Jia}
will work toward the Master’s degree at the College of Computer Science, Nankai University, Tianjin, China. His research interests include computer vision and pattern recognition.
\end{IEEEbiographynophoto}
\vspace{-30pt}

\begin{IEEEbiographynophoto}{Guiguang Ding}
received his Ph.D.\ degree from Xidian University, China, in 2004. He is currently an associate professor of School of Software, Tsinghua University. His current research centers on the area of multimedia information retrieval, computer vision and machine learning. %He has published over 80 scientific papers in major journals and conferences, including the IEEE TIP, TMM, TKDE, SIG IR, AAAI, ICML, IJCAI, CVPR, ICCV, \emph{etc}.
\end{IEEEbiographynophoto}
\vspace{-30pt}

\begin{IEEEbiographynophoto}{Tat-Seng Chua}
joined the National University of Singapore, Singapore, in 1983, and spent three years as a Research Staff  Member with the Institute of Systems Science, National University of Singapore. He was the Acting and Founding Dean of the School of Computing, National University of Singapore, from 1998 to 2000. He is currently the KITHCT Chair Professor with the School of Computing, National University of Singapore. His research interests include multimedia information retrieval, multimedia question answering, and the analysis and structuring of user-generated contents.
%Dr.\ Chua has organized and served as a Program Committee Member of numerous international conferences in the areas of computer graphics, multimedia, and text processing. He was the Conference Co-Chair of ACM Multimedia in 2005, the Conference on Image and Video Retrieval in 2005 and ACM SIGIR in 2008, and the Technical PC Co-Chair of SIGIR in 2010. He serves on the editorial boards of the ACM Transactions of Information Systems, Foundation and Trends in Information Retrieval, The Visual Computer, and Multimedia Tools and Applications. He is on the Steering Committees of the International Conference on Multimedia Retrieval, Computer Graphics International, and Multimedia Modeling Conference Series. He serves as a member of international review panels of two large-scale research projects in Europe. He is the Independent Director of two listed companies in Singapore.
\end{IEEEbiographynophoto}
\vspace{-30pt}

\begin{IEEEbiographynophoto}{Bj{\"o}rn W. Schuller}
heads Imperial College London's Group on Language Audio \& Music (GLAM), is CEO of the Audio Intelligence company audEERING, and a Full Professor at University of Augsburg/Germany in Computer Science. He received his diploma, doctoral, and habilitation degrees from TUM in Munich/Germany in EE/IT. Previous positions of his include Full Professor at the University of Passau/Germany and Visiting Professor, Associate, and Scientist at VGTU/Lithuania, University of Geneva/Switzerland, Joanneum Research/Austria, Marche Polytechnic University/Italy, and CNRS-LIMSI/France. His technical publications focus on machine intelligence for affective multimedia analysis. He is a Fellow of the IEEE, President-Emeritus of the AAAC, and the Editor in Chief of the IEEE TAFFC.
%, Associate Editor of the IEEE TNNLS, the IEEE TCYB, and the IEEE SPL among other Associate and multiple Guest Editorships, and a General Chair of the oncoming IEEE ACII 2019 and Technical Chair of Interspeech 2019 among various past according and further roles. He received a range of awards including being honoured as one of 40 extraordinary scientists under the age of 40 by the World Economic Forum in 2015. In 2017, his company secured the 1st place as ``Innovator of The Year'' of the Digital Marketing Innovation World Cup. His research has garnered over 10 million USD in extramural funding. Advisory board activities comprise his role as consultant of global enterprises such as BARCLAYS, GN, HUAWEI and SAMSUNG.
\end{IEEEbiographynophoto}
\vspace{-30pt}

\begin{IEEEbiographynophoto}{Kurt Keutzer}
(F'96) received his Ph.D.\ degree in Computer Science from Indiana University in 1984 and then joined the research division of AT\&T Bell Laboratories. In 1991 he joined Synopsys, Inc.\ where he ultimately became Chief Technical Officer and Senior Vice-President of Research. In 1998, Kurt became Professor of Electrical Engineering and Computer Science at the University of California at Berkeley. Kurt's research group is currently focused on using parallelism to accelerate the training and deployment of Deep Neural Networks for applications in computer vision, speech recognition, multi-media analysis, and computational finance.
%Kurt has published six books, over 250 refereed articles, and is among the most highly cited authors in Hardware and Design Automation.
Kurt is a Life Fellow of the IEEE.
\end{IEEEbiographynophoto}

\end{document}